\documentclass{article}
\usepackage{spconf,,graphicx}
\usepackage{caption}
\usepackage[table,xcdraw,dvipsnames]{xcolor}
 \usepackage{booktabs}
\usepackage{algorithm, setspace}
\usepackage{algpseudocode}
\usepackage{url}

\newcolumntype{P}[1]{>{\centering\arraybackslash}p{#1}}
\newcommand{\shortname}{SimSAM}
\usepackage{amsmath}
\newcommand{\beginsupplement}{%
        \setcounter{table}{0}
        \renewcommand{\thetable}{S\arabic{table}}%
        \setcounter{figure}{0}
        \renewcommand{\thefigure}{S\arabic{figure}}%
     }

\title{SimSAM: Simple Siamese Representations Based Semantic Affinity Matrix for Unsupervised Image Segmentation}
\name{Author(s) Name(s)\thanks{Thanks to XYZ agency for funding.}}
\address{Author Affiliation(s)}

\name{Chanda Grover Kamra$^{\star}$ \qquad Indra Deep Mastan$^{\dagger}$ \qquad Nitin Kumar$^{\wedge}$ \qquad Debayan Gupta$^{\star}$}  
  \address{$^{\star}$ Ashoka University - Sonipat, India. $^{\dagger}$ Indian Institute of Technology (IIT-BHU) Varanasi, India.\\
  $^{\wedge}$ Shiv Nadar University - New Delhi, India.
      }

\begin{document}
\maketitle

\begin{abstract}
Recent developments in self-supervised learning (SSL) have made it possible to learn data representations without the need for annotations.
%, which enhances the performance of various downstream tasks. 
Inspired by the non-contrastive SSL approach (SimSiam), we introduce a novel framework \shortname~to compute the Semantic Affinity Matrix, which is significant for unsupervised image segmentation. Given an image, \shortname~first extracts features using pre-trained DINO-ViT, then projects the features to predict the correlations of dense features in a non-contrastive way. We show applications of the Semantic Affinity Matrix in object segmentation and semantic segmentation tasks. Our code is available at https://github.com/chandagrover/SimSAM. 
\end{abstract}


\section{Introduction}
\label{sec:intro}
Self-supervised learning (SSL) based methods \cite{mukhoti2023open, caron2021emerging, chen2020simple, byol_nips2020, zbontar2021barlow} have shown promising results by learning generic visual representations without requiring labeled data. Self-supervised approaches can be broadly classified into contrastive   \cite{chen2020simple, mukhoti2023open} and non-contrastive~\cite{byol_nips2020, chen2021exploring, zbontar2021barlow, bardes2022vicreg, caron2021emerging} approaches. Contrastive approaches use positive and negative examples, while non-contrastive approaches rely only on positive samples to learn representations. One of the 
goals of self-supervised learning is to learn instance-based discriminative features for various tasks in the vision domain like classification, object segmentation \cite{van2021unsupervised, melas2022deep} and semantic segmentation \cite{Li_2023_CVPR, zadaianchuk2023unsupervised, mukhoti2023open} etc.  
\begin{figure}  
  \begin{minipage}{0.12\linewidth}
     \centering
     \scriptsize
        a) Input Image
    \end{minipage}
   \begin{minipage}{0.19\linewidth}
     \centering
     \scriptsize
        b) Deep Cut \cite{Aflalo_2023_ICCV}
    \end{minipage}
    \begin{minipage}{0.19\linewidth}
    \centering 
    \scriptsize
        c) DSM \cite{melas2022deep}
    \end{minipage}
    \begin{minipage}{0.19\linewidth}
     \centering
     \scriptsize
       \textbf{d) Ours}
    \end{minipage}
    \begin{minipage}{0.19\linewidth}
     \centering
     \scriptsize
        e) Ground Truth
    \end{minipage}
   \begin{minipage}{0.19\linewidth}
     \centering             
     \includegraphics[width=0.99\linewidth]{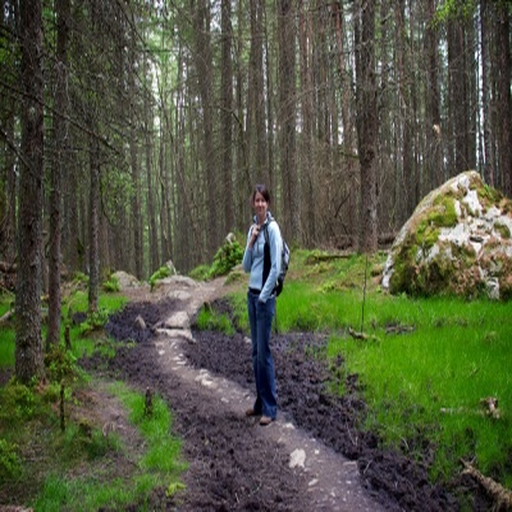}
    \end{minipage}
   \begin{minipage}{0.19\linewidth}
     \centering             
     \includegraphics[width=0.99\linewidth]{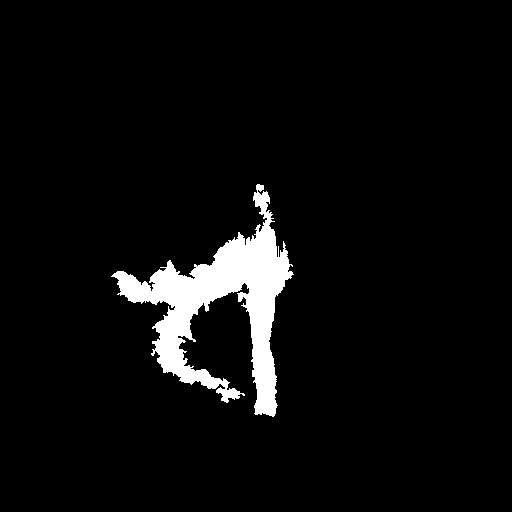}
    \end{minipage}
    \begin{minipage}{0.19\linewidth}
     \centering             
     \includegraphics[width=0.99\linewidth]{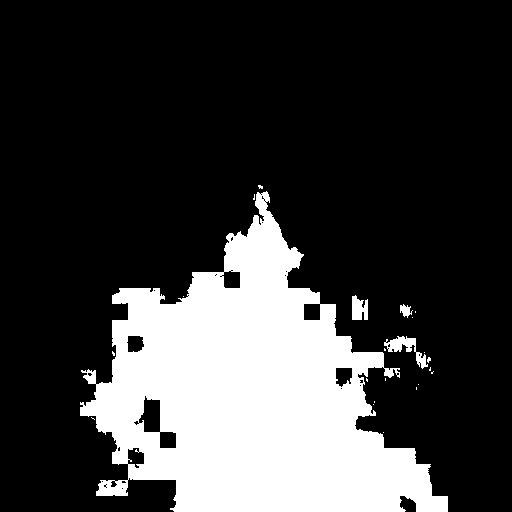}
    \end{minipage}
    \begin{minipage}{0.19\linewidth}
     \centering   
    \includegraphics[width=0.99\linewidth]{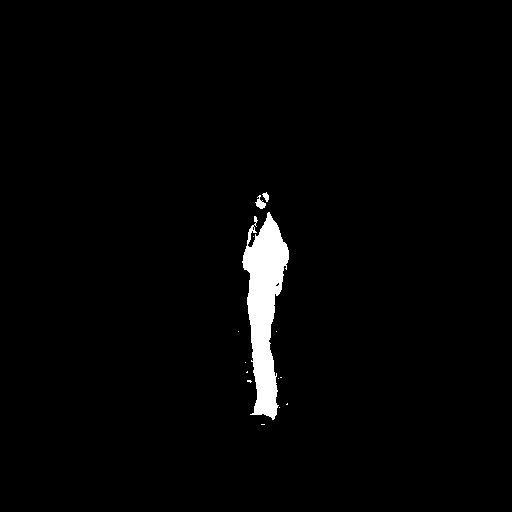}
    \end{minipage}
   \begin{minipage}{0.19\linewidth}
     \centering             
     \includegraphics[width=0.99\linewidth]{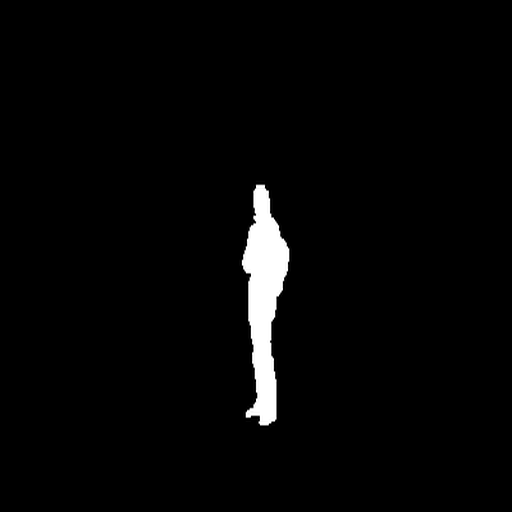}
    \end{minipage}
   \vspace{0.2cm}     
   \hrule
   \vspace{0.2cm}
\begin{minipage}{0.24\linewidth}
     \centering
     \scriptsize
       f) Input Image
    \end{minipage}
    \begin{minipage}{0.24\linewidth}
    \centering
    \scriptsize
    g) DSM \cite{melas2022deep}
    \end{minipage}
    \begin{minipage}{0.24\linewidth}
     \centering
     \scriptsize
       \textbf{h) Ours}
    \end{minipage}
    \begin{minipage}{0.24\linewidth}
     \centering
     \scriptsize
    i) Ground Truth
    \end{minipage}
    \begin{minipage}{0.24\linewidth}
     \centering        
     \includegraphics[width=0.99\linewidth]{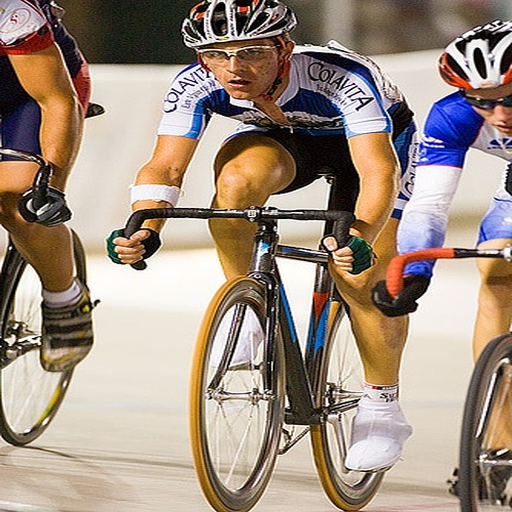}
    \end{minipage}
     \begin{minipage}{0.24\linewidth}
     \centering        \includegraphics[width=0.99\linewidth]{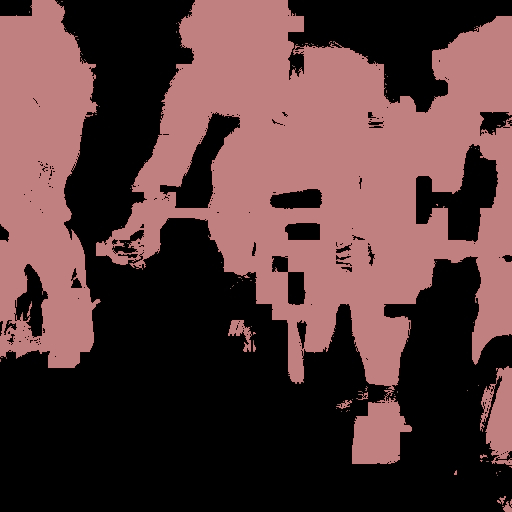}
    \end{minipage}
    \begin{minipage}{0.24\linewidth}
     \centering
        \includegraphics[width=0.99\linewidth]{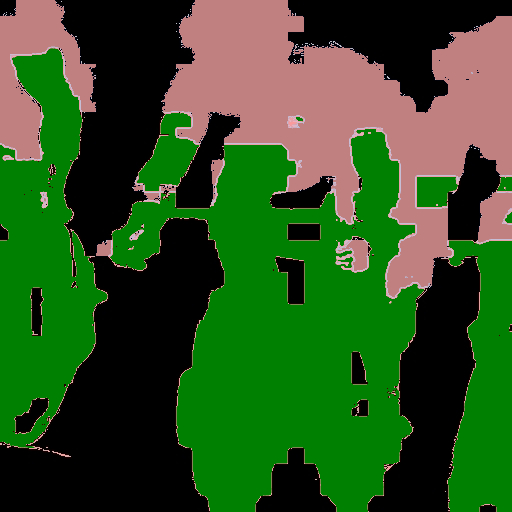}
    \end{minipage}
    \begin{minipage}{0.24\linewidth}
     \centering
        \includegraphics[width=0.99\linewidth]{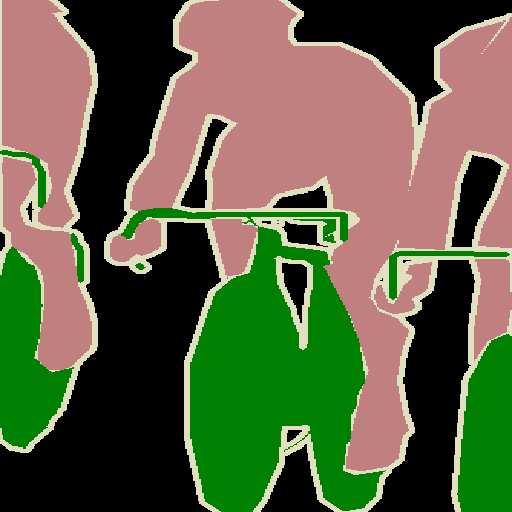}
    \end{minipage}
    \caption{Top row shows the qualitative results of object segmentation (a-e), and the bottom row presents the results of semantic segmentation (f-i).}
\label{fig:mainresultfig}
\end{figure}

Recently, Deep Spectral Methods (DSM) \cite{melas2022deep} showed a strong baseline of unsupervised segmentation utilizing the semantic affinity matrix computed from DINO-ViT \cite{caron2021emerging}. Mark et al.  \cite{hamilton2022unsupervised} empirically showed that pre-trained DINO-ViT \cite{caron2021emerging} dense feature correlations are semantically consistent. However, the affinity matrix in  DSM \cite{melas2022deep} is computed by taking simply the dot product of the extracted deep features, with an assumption on the alignment of the deep features. We take inspiration from a non-contrastive SSL approach for learning the Semantic Affinity Matrix, which is important for unsupervised image segmentation.

% Inspired by SimSiam \cite{chen2021exploring}, we model the computation of affinity matrix by projecting deep features into an embedding space where the prediction of affinities is carried out accurately.
% The  Semantic Affinity Matrix of deep features extracted from an image is essential to conceptualize the objects present in the image.

Fig.~\ref{fig:mainresultfig} shows two challenging scenarios of feature correlations on object segmentation and semantic segmentation tasks. The top row shows that the visual features of the man's lower half are similar to the road. Similarly, in the bottom row, the existing method DSM \cite{melas2022deep} conflates the sandy tyre features of a bicycle with the hands of a person riding the bicycle. In such scenarios (Fig.~\ref{fig:mainresultfig}), feature affinity computation using dense features - as in DSM \cite{melas2022deep}- does not preserve useful semantics (Ablation Study-I in Sec.~\ref{sec:Ablation_Study}). Non-preservation of essential semantics is indicated by lower Frobenious, mIoU and accuracy scores for randomly sampled images from the ECSSD dataset.
% We study feature affinity computation, inspired by \cite{melas2022deep, hamilton2022unsupervised}, which is essential for producing better spectral segmentation methods and hence stylized outputs.

\begin{figure*}[!ht]
\includegraphics[width=0.99\textwidth]{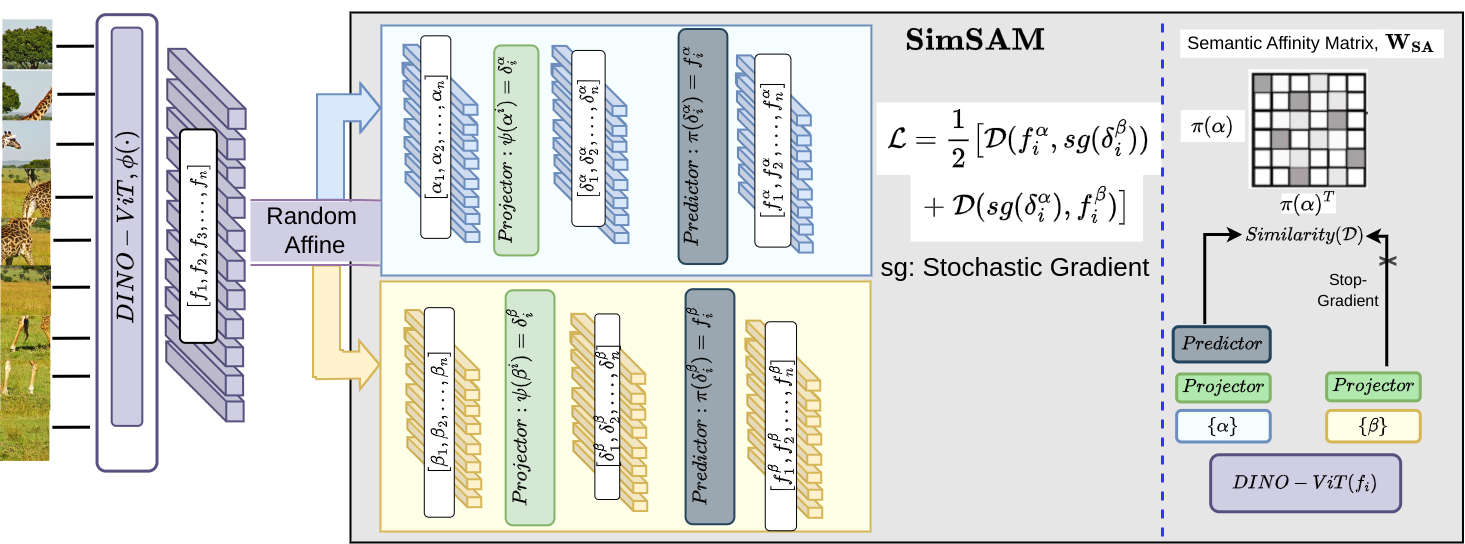}
\caption{The left side illustrates the \shortname~framework. The right side provides an overview of the computation of the semantic affinity matrix $W_{SA}$. First, we extract features using DINO-ViT. Then, two views ($\alpha_i$ and $\beta_i$) are obtained with Random Affine transformations. These views are processed by projector network $\psi$ (consists of a non-linear layer),  followed by a predictor $\pi$ (consists of a linear layer), and loss $\mathcal{L}$ is minimized to train the projector and predictor (with stop-grad). Finally, $W_{SA}$ is computed for spectral segmentation.}
\label{fig:featAffinity_DSCS}
\end{figure*}

In this work, we propose \shortname~ framework that is semantically consistent and computes a dense feature affinity matrix (Sec.~\ref{sec:our_approach} and Table~\ref{Table:image_seg}). Our key idea is to use the discriminative representations learned by Simple Siamese Network (SimSiam) \cite{chen2021exploring}  in the non-contrastive setting to compute the feature affinity matrix for DSM \cite{melas2022deep} segmentation. \shortname~ uses only the dense representations extracted from pre-trained DINO-ViT \cite{caron2021emerging}. 
% We use symmetric loss with stop-grad inspired by the SimSiam \cite{chen2021exploring} framework to train the projector and predictor. 
Finally, we compute the semantic affinity matrix for deep spectral segmentation (see Fig.~\ref{fig:featAffinity_DSCS}). We observe that incorporating Siamese-based architecture results in learning better feature correlations for similar and dissimilar color regions of an image, with better qualitative and quantitative results (see Fig.~\ref{fig:masks} and Table \ref{Table:image_seg}). 

Our contributions are summarized below: 
\begin{enumerate}
    \setlength{\itemsep}{1pt}
     \item We propose \shortname~framework (Algorithm~\ref{algo:feat_affinity}) to compute dense feature correlations, which are essential to preserve spectral segmentation semantics.  (Sec.~\ref{sec:our_approach}). 
    % \item Our framework outperforms in mIoU scores vs the baseline methods on object segmentation (Table~\ref{Table:image_seg}). 
    \item We perform extensive experiments to validate that the affinity matrix computed from \shortname~leads to semantically consistent segmentations. (See Sec.~\ref{sec:Ablation_Study}.)    
\end{enumerate}
\section{Related Work}
\label{sec:related_work}
\noindent\textbf{Self-Supervised Learning.} 
Recently, self-supervised learning \cite{bardes2022vicreg, mukhoti2023open, li2021efficient}, has made notable advancements. Most recent methods primarily adopt the joint-embedding framework utilizing a siamese network architecture \cite{bromley1993signature}. These approaches are generally categorized into two groups: contrastive and non-contrastive methods.
Contrastive \cite{bromley1993signature, chen2020simple} methods explicitly push the embeddings of dissimilar images away from each other and hence tend to be costly. While, non-contrastive methods \cite{byol_nips2020, chen2021exploring} do not rely on contrastive samples but employ several tricks like stop-gradient operation, and momentum encoder to avoid the need for negative samples.

\noindent\textbf{Unsupervised Segmentation.}
 Based on SSL, various approaches have been proposed for unsupervised segmentation \cite{van2021unsupervised, Li_2023_CVPR, zadaianchuk2023unsupervised, kirillov2023segment}. Self-supervised ViTs \cite{image2021iclr} have shown the ability to represent pixel-level semantic relationships. Caron et al. \cite{caron2021emerging} showed that DINO’s class attention produces semantically meaningful and localized salient object segmentation. Li et al. \cite{Li_2023_CVPR} address the challenge of over/under-clustering within DINO representations in images to extract fundamental concepts. 
 % Comus \cite{zadaianchuk2023unsupervised} introduced a methodology that utilizes unsupervised saliency masks along with self-supervised feature clustering. 
 Aflalo et al. \cite{Aflalo_2023_ICCV} proposed lightweight graph neural networks and optimized correlation-clustering-based objective functions for segmentation.
% However, their proposed approach required the sensitivity parameter $\alpha$ for regulating the number of clusters.
 Melas et al. \cite{melas2022deep} proposed the deep spectral method for segmentation on the dense features extracted from  DINO \cite{caron2021emerging} features. 
\section{Our Approach}
\label{sec:our_approach}
The \shortname~framework is shown in Fig.~\ref{fig:featAffinity_DSCS} and Algorithm~\ref{algo:feat_affinity}. The aim is to compute the semantic affinity matrix to improve unsupervised image segmentation. We describe \shortname~ framework as follows:- \\~

%Our framework, \shortname~(Fig.~\ref{fig:featAffinity_DSCS} and Algorithm~\ref{algo:feat_affinity}) is aimed at computing the semantic affinity matrix to improve unsupervised image segmentation. The steps for our Algorithm~\ref{algo:feat_affinity} are as follows:
 
\noindent \textbf{Feature Extraction.}
The input image $I \in R^{3 \times M \times N}$ is passed through pre-trained DINO-ViT  \cite{image2021iclr} $\phi$ to extract dense features $\phi(I) = \left[ F_1, F_2,\cdots, F_L\right]$, where $L$ is the total number of layers. These features are in the form of queries, keys, and values. Similar to \cite{melas2022deep}, we observe that the features from the keys K of the last attention layer $F_L^K = \left[ f_1, f_2, \cdots f_n \right]$ are useful to be used as the patch representations for segmentation. Where the number of patches $n = \frac{M N}{P^2 }$ and $P \times P$ represent the patch size. We consider $F_L^K$ features in SimSAM~ framework for semantic affinity computation and compute vanilla affinity matrix $W_{A}$, simply as the normalized matrix as follows: 
\begin{equation}
\label{eq:5_2}
    W_{A}= (F_L^K)(F_L^K)^T - \mathcal{M}  \big( (F_L^K)(F_L^K)^T \big)
\end{equation}
Here, the normalization is done by subtracting the mean ($\mathcal{M}$) of affinity value $\mathcal{M}  \big( (F_L^K)(F_L^K)^T \big)$  from the correlation matrix $ (F_L^K)(F_L^K)^T$. 
\begin{algorithm*}[!ht] \setstretch{1.22}
    \begin{algorithmic}[1]
    \caption{Learning semantic affinity matrix for SimSAM.}
    \label{algo:feat_affinity}
     \Statex \hspace*{-0.5cm} \textbf{Input:} Input Image $I$, Dino-ViT $\phi$, Projector $\psi$, Predictor $\pi$.  
     \State $\phi(I) = \big[ F_1, F_2 \cdots, F_L \big]$   \hspace{1cm}\hfill {\color{gray} \textit{Feature layers of network $\phi(\cdot)$ ~~~~}} 
     \State $W_{A}$ = $(F_L^K)(F_L^K)^T$ - $\mathcal{M}  \big( (F_L^K)(F_L^K)^T \big)$ \hspace{1cm}\hfill {\color{gray} \textit{Vanilla Affinity Matrix ~~~~}}
     \State $\alpha$, $\beta$ = $RA(F_L^K)$ = $RA(\big[ f_1, f_2 \cdots, f_n \big])$
     \hspace{1cm}\hfill {\color{gray} \textit{ Random Affine on keys of last attention layer ~~~~}} 
    \State $\delta^\alpha$ = $\big[ \delta^\alpha_1,  \delta^\alpha_2, \ldots, \delta^\alpha_n\big]$, where $\delta^\alpha_i = \psi(\alpha_i)$ \hspace{1cm}\hfill {\color{gray} \textit{ MLP Non-Linear Projection on $\alpha$~~~~}} 
      \State $\delta^\beta$ = $\big[ \delta^\beta_1, \delta^\beta_2, \ldots, \delta^\beta_n\big]$, where $\delta^\beta_i = \psi(\beta_i)$ \hspace{1cm}\hfill {\color{gray} \textit{ MLP Non-Linear Projection on $\beta$~~~~}}
    \State  $f^\alpha  = \big[f^\alpha_1,  f^\alpha_2, \ldots, f^\alpha_n) \big]$, where $f^\alpha_i = \pi(\delta^\alpha_i)$  \hspace{1cm}\hfill {\color{gray} \textit{semantic features from $\alpha$ ~~~~}}
     \State $f^\beta = \big[f^\beta_1, f^\beta_2, \ldots, f^\beta_n \big]$, where  $f^\beta_i = \pi(\delta^\beta_i)$  \hspace{1cm}\hfill {\color{gray} \textit{ semantic features from $\beta$ ~~~~}}
     \State $\mathcal{L}$ = $\frac{1}{2}\big( \mathcal{D}(f^\alpha_i,sg(\delta^\beta_i)) + \mathcal{D}(sg(\delta^\alpha_i),f^\beta_i) \big)$  \hspace{1cm}\hfill {\color{gray} \textit{ Predictor Network and loss function ~~~~}} 
    \State $W_{SA}$ = $\big[\pi ( \alpha_1), \cdots, \pi (\alpha_n) \big] \big[\pi (\alpha_1), \cdots, \pi( \alpha_n) \big]^T$  \hspace{1cm}\hfill {\color{gray} \textit{ semantic affinity matrix ~~~~}}  
     \State $W_{feat} = W_{A} + \kappa ~  W_{SA}$  \hspace{1cm}\hfill {\color{gray} \textit{ Final Semantic Matrix ~~~~}} 
    \end{algorithmic}
\end{algorithm*}

\noindent \textbf{Random Affine Transformation.} We transform the extracted features (tokens) into two views $\alpha$ and $\beta$ through random affine $RA(\cdot)$ transformation: 
\begin{equation}
% \begin{split}
        \alpha, \beta = RA(\big[ f_1, f_2, \ldots, f_n\big]) \\
            % & = \big[ \beta_1, \beta_2, \beta_3, \ldots, \beta_n\big]
            % \end{split}
\end{equation}
Here, $ \alpha = \big[ \alpha_1, \alpha_2, \ldots, \alpha_n\big]$ and $ \beta = \big[ \beta_1, \beta_2, \ldots, \beta_n\big]$ are the corresponding tokens of views $\alpha$ and $\beta$, respectively. The random affine transformation consists of scaling, translation, rotation, etc. In our non-contrastive learning framework, only the extracted features (positive pairs) are used for learning.

\noindent \textbf{Projector. } Our \shortname~framework leverages SimSiam \cite{chen2021exploring} in two possible ways: i) We operate on Dino-ViT pre-trained feature space, while SimSiam operates on the raw image space. ii) Our projector comprises a single non-linear layer in its projection head followed by an ELU activation function and the predictor is a single linear layer. SimSiam's projector consists of 3 layers in their projection head with ReLU activation function, and their predictor comprises two layers, again with ReLU. Further, their output layer does not apply any activation function.

% The projector comprises a parametric non-linear projector layer; the predictor has a single linear layer. 
We conduct ablation studies with simple and complex configurations for the projector and predictor (See Fig.~\ref{Fig:featuremap_ablation}). Projector processes the two views ($\alpha$ and $\beta$). MLP projection ($\psi$) head at the projector shares the weights between two views of token features and then projects them as follows:
\begin{equation}
     \delta^\alpha = \big[\psi(\alpha_1), \psi(\alpha_2),\cdots, \psi(\alpha_n)\big] = \big[ \delta^\alpha_1, \delta^\alpha_2, \ldots, \alpha^\alpha_n \big] 
\end{equation} 
\begin{equation}
     \delta^\beta = \big[\psi(\beta_1), \psi(\beta_2),\cdots, \psi(\beta_n)\big] = \big[ \delta^\beta_1, \delta^\beta_2, \ldots, \delta^\beta_n \big] 
\end{equation}    
Here, $\delta^\alpha_i$ and $\delta^\beta_i$ are of the same dimensions as $\alpha_i$ and $\beta_i$. Therefore, the MLP projection head projects the dense features into a space of similar dimensions.

\noindent \textbf{Predictor.} SimSAM framework uses an MLP as predictor $\pi$ to predict the embedding that preserves semantics when computing the affinity matrix. $\pi$ is a simple dense layer without activation. We describe how we apply predictor $\pi$ as below: 
\begin{equation}
    f^\alpha = \big[f^\alpha_1, \ldots, f^\alpha_n) \big] = \big[\pi(\delta_1^{\alpha}), \ldots, \pi(\delta_n^{\alpha}) \big]
\end{equation}
\begin{equation}
    f^\beta = \big[f^\beta_1, \ldots, f^\beta_n \big] = \big[\pi(\delta_1^{\beta}), \ldots, \pi(\delta_n^{\beta}) \big]
\end{equation}
$f^\alpha$ and $f^\beta$ are learned semantic features from views $\alpha$ and $\beta$. The dimensions of the predicted dense features $\pi(\delta_i^{\alpha})$ and $\pi(\delta_i^{\beta})$ are same as its inputs $\delta_i^{\alpha}$ and $\delta_i^{\beta}$ respectively.

\begin{figure*}[!ht]
    % \centering
    \begin{minipage}{0.12\linewidth}
     \centering
        Input Image
    \end{minipage}
   \begin{minipage}{0.12\linewidth}
     \centering
        Deep Cut \cite{Aflalo_2023_ICCV}
    \end{minipage}
    \begin{minipage}{0.12\linewidth}
    \centering
        DSM \cite{melas2022deep}
    \end{minipage}
        \begin{minipage}{0.12\linewidth}
     \centering
     \small
       SimSAM~(Ours)
    \end{minipage}
    \begin{minipage}{0.12\linewidth}
     \centering
        Input Image
    \end{minipage}
   \begin{minipage}{0.12\linewidth}
     \centering
        Deep Cut \cite{Aflalo_2023_ICCV}
    \end{minipage}
    \begin{minipage}{0.12\linewidth}
    \centering
        DSM \cite{melas2022deep}
    \end{minipage}
    \begin{minipage}{0.12\linewidth}
     \centering
     \small
       SimSAM~(Ours)
    \end{minipage}
    \begin{minipage}{0.12\linewidth}
     \centering             
     \includegraphics[width=0.99\linewidth]{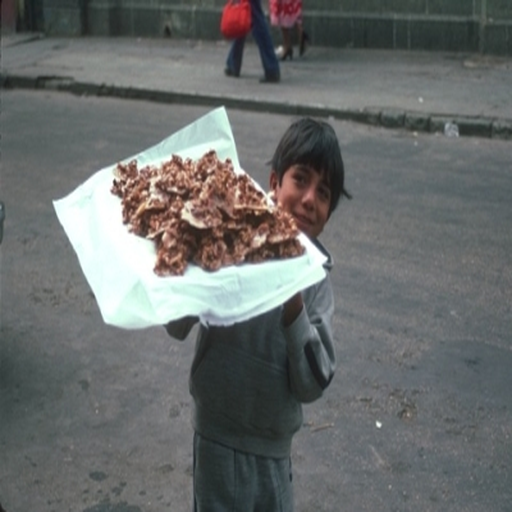}\llap{\raisebox{1.35cm}{%  move next graphics to top right corner
      \includegraphics[height=0.7cm]{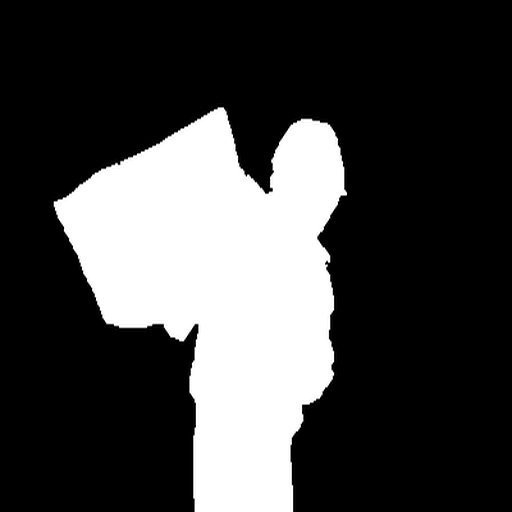}%
    }}
    \end{minipage}
    % \hspace{-0.2cm}
   \begin{minipage}{0.12\linewidth}
     \centering             
     \includegraphics[width=0.99\linewidth]{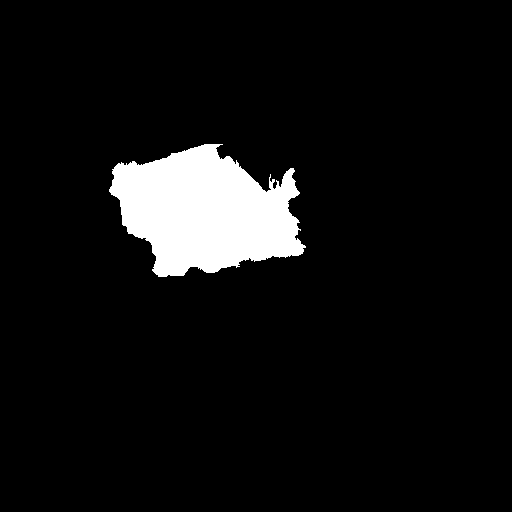}
    \end{minipage} 
    \begin{minipage}{0.12\linewidth}
     \centering   
     \includegraphics[width=0.99\linewidth]{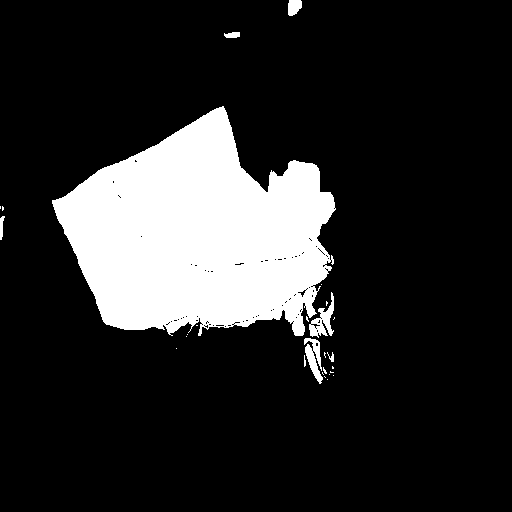}
    \end{minipage}
    \begin{minipage}{0.12\linewidth}
     \centering             
     \includegraphics[width=0.99\linewidth]{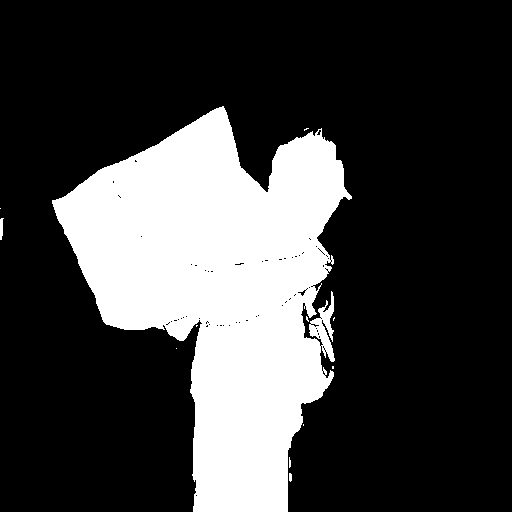}
    \end{minipage}    
    \hfill
    \textbf{\vline}
    \hfill
    \begin{minipage}{0.12\linewidth}
     \centering             
     \includegraphics[width=0.99\linewidth]{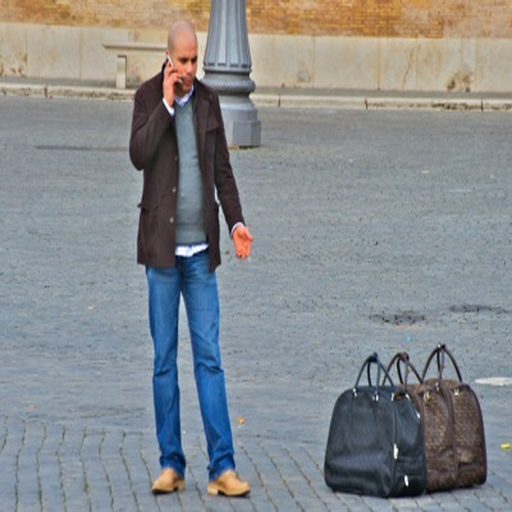}\llap{\raisebox{1.35cm}{%  move next graphics to top right corner
      \includegraphics[height=0.7cm]{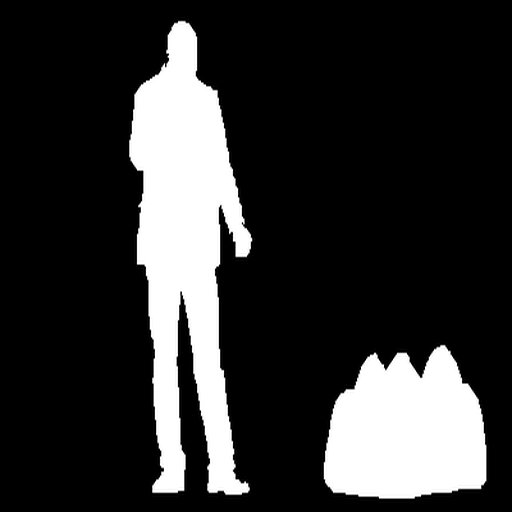}%
    }}
    \end{minipage}
    % \hspace{-0.2cm}
   \begin{minipage}{0.12\linewidth}
     \centering             
     \includegraphics[width=0.99\linewidth]{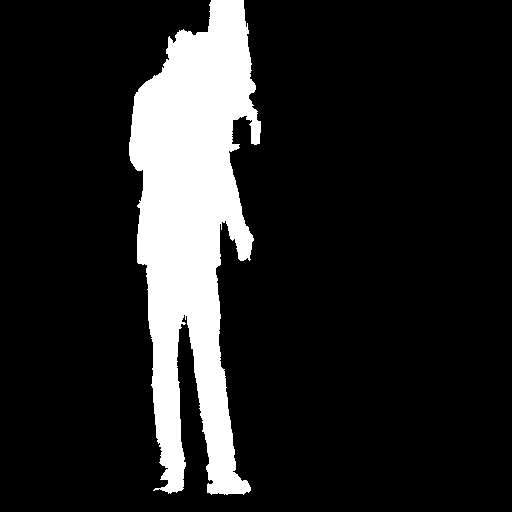}
    \end{minipage}
    \begin{minipage}{0.12\linewidth}
     \centering            \includegraphics[width=0.99\linewidth]{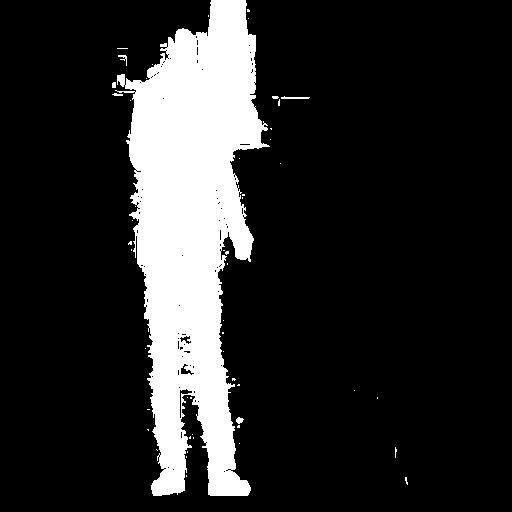}
    \end{minipage}
    \begin{minipage}{0.12\linewidth}
     \centering             \includegraphics[width=0.99\linewidth]{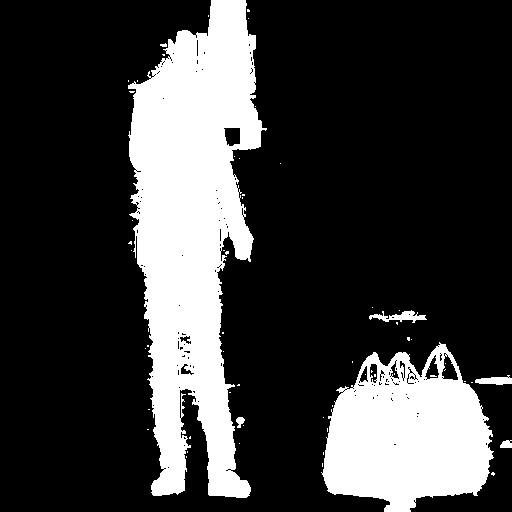}
    \end{minipage}
   % \begin{minipage}{0.12\linewidth}
   %   \centering             \includegraphics[width=0.99\linewidth]{images/Masks/0122_resized.png}\llap{\raisebox{1.35cm}{%  move next graphics to top right corner
   %    \includegraphics[height=0.7cm]{images/Masks/0122_resized_ground.png}%
   %  }}
   %  \end{minipage}
   %  % \hspace{-0.2cm}
   % \begin{minipage}{0.12\linewidth}
   %   \centering             
   %   \includegraphics[width=0.99\linewidth]{images/Masks/0122_resized_deepcut.png}
   %  \end{minipage}
   %  \begin{minipage}{0.12\linewidth}
   %   \centering            \includegraphics[width=0.99\linewidth]{images/Masks/0122_resized_srg.png}
   %  \end{minipage}
   %  \begin{minipage}{0.12\linewidth}
   %   \centering            \includegraphics[width=0.99\linewidth]{images/Masks/0122_resized_ours.png}
   %  \end{minipage}
    \begin{minipage}{0.12\linewidth}
     \centering             
     \includegraphics[width=0.99\linewidth]{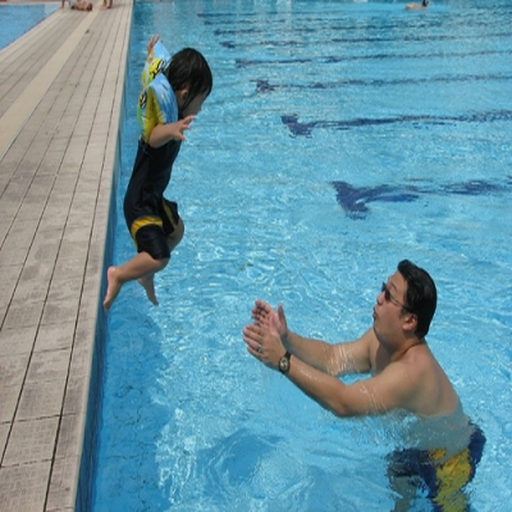}\llap{\raisebox{1.35cm}{%  move next graphics to top right corner
      \includegraphics[height=0.7cm]{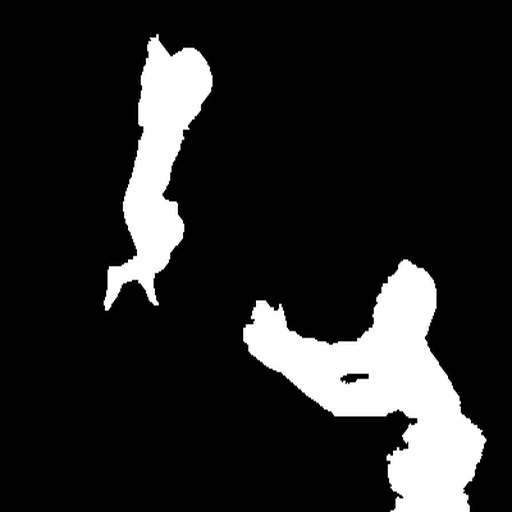}%
    }}
    \end{minipage}
    % \hspace{-0.2cm}
   \begin{minipage}{0.12\linewidth}
     \centering             
     \includegraphics[width=0.99\linewidth]{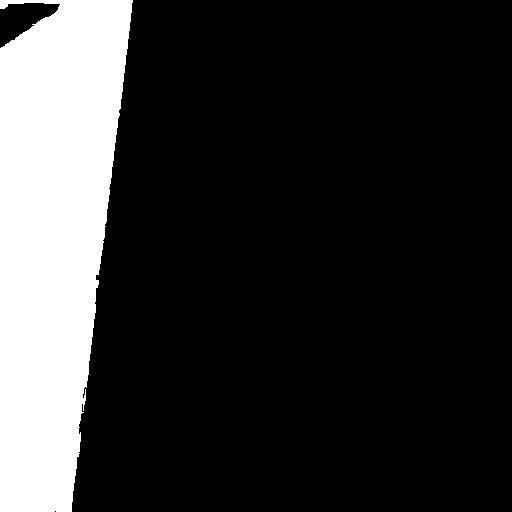}
    \end{minipage}
    \begin{minipage}{0.12\linewidth}
     \centering            \includegraphics[width=0.99\linewidth]{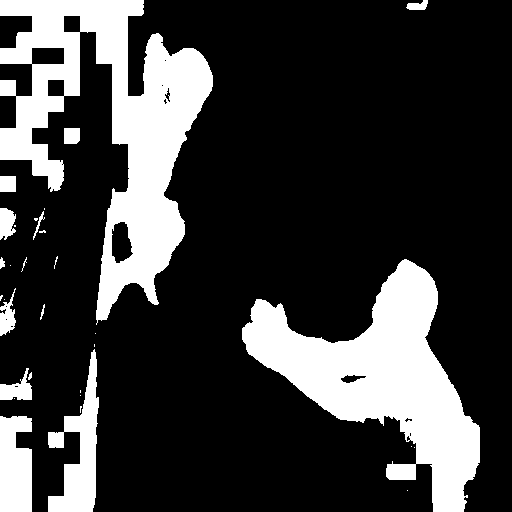}
    \end{minipage}
    \begin{minipage}{0.12\linewidth}
     \centering             \includegraphics[width=0.99\linewidth]{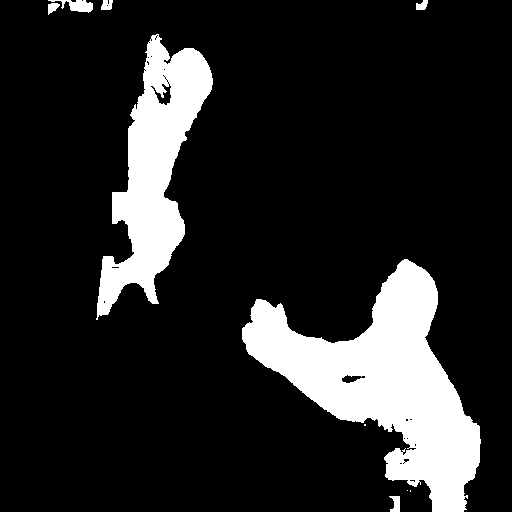}
    \end{minipage} 
    \hfill
    \textbf{\vline}
    \hfill
     \begin{minipage}{0.12\linewidth}
     \centering            
     \includegraphics[width=0.99\linewidth]{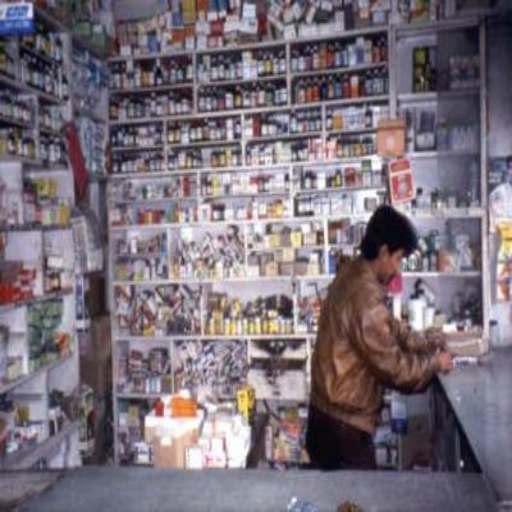}\llap{\raisebox{1.35cm}{%  move next graphics to top right corner
      \includegraphics[height=0.7cm]{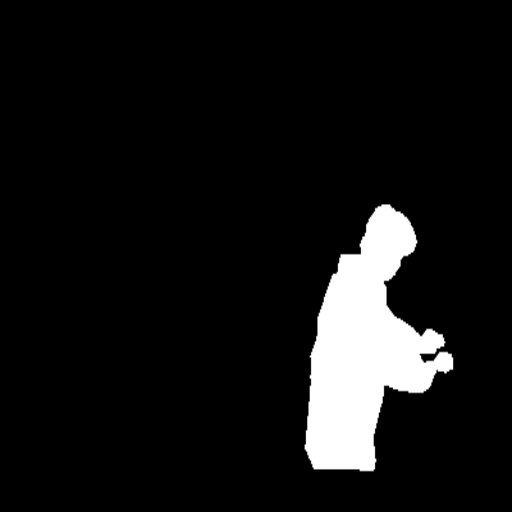}%
    }}
    \end{minipage}
    % \hspace{-0.2cm}
   \begin{minipage}{0.12\linewidth}
     \centering             
     \includegraphics[width=0.99\linewidth]{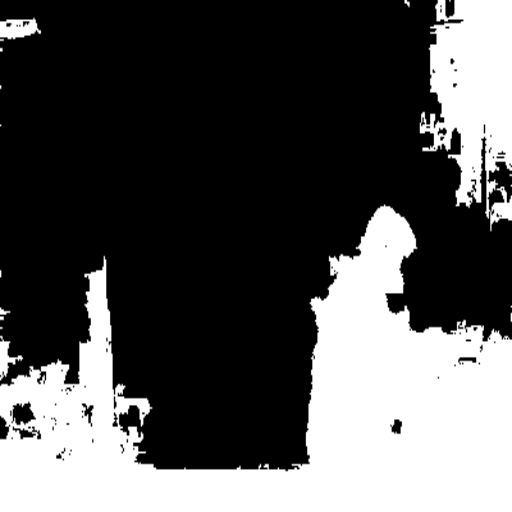}
    \end{minipage}
    \begin{minipage}{0.12\linewidth}
     \centering            \includegraphics[width=0.99\linewidth]{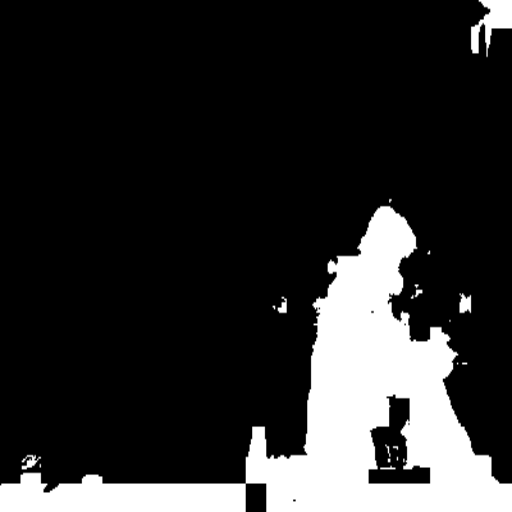}
    \end{minipage}
    \begin{minipage}{0.12\linewidth}
     \centering             \includegraphics[width=0.99\linewidth]{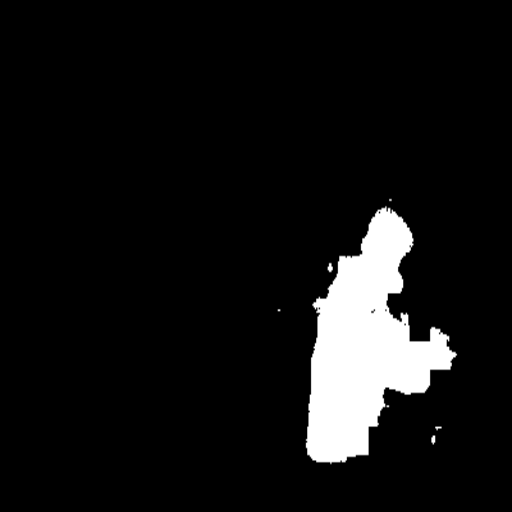}
    \end{minipage}
   %  \begin{minipage}{0.12\linewidth}
   %   \centering             
   %   \includegraphics[width=0.99\linewidth]{images/Masks/0511_resized.png}\llap{\raisebox{1.35cm}{%  move next graphics to top right corner
   %    \includegraphics[height=0.7cm]{images/Masks/0511_resized_ground.png}%
   %  }}
   %  \end{minipage}
   %  % \hspace{-0.2cm}
   % \begin{minipage}{0.12\linewidth}
   %   \centering             
   %   \includegraphics[width=0.99\linewidth]{images/Masks/0511_resized_deepcut.png}
   %  \end{minipage}
   %  \begin{minipage}{0.12\linewidth}
   %   \centering            \includegraphics[width=0.99\linewidth]{images/Masks/0511_resized_srg.png}
   %  \end{minipage}
   %  \begin{minipage}{0.12\linewidth}
   %   \centering             \includegraphics[width=0.99\linewidth]{images/Masks/0511_resized_ours.png}
   %  \end{minipage}   
    \begin{minipage}{0.12\linewidth}
     \centering             
     \includegraphics[width=0.99\linewidth]{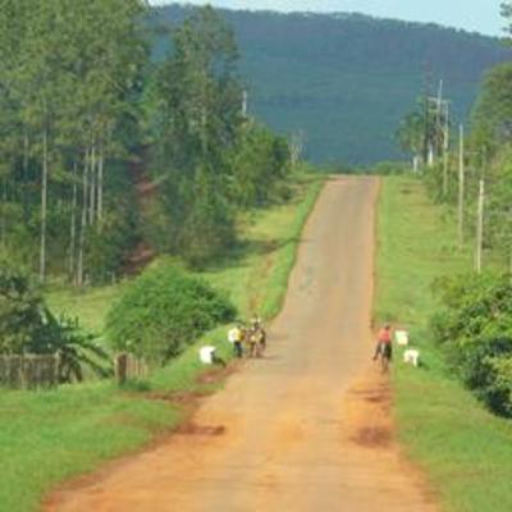}\llap{\raisebox{1.35cm}{%  move next graphics to top right corner
      \includegraphics[height=0.7cm]{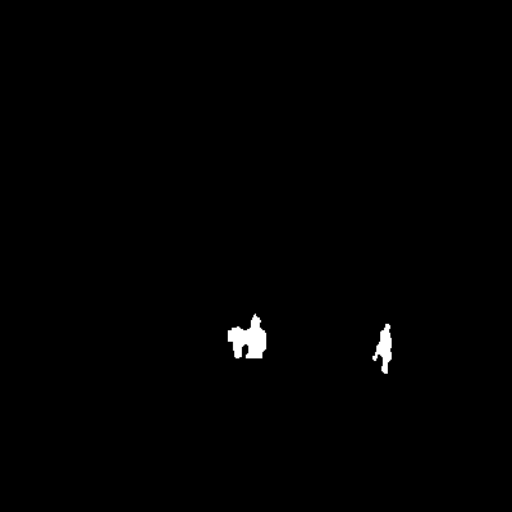}%
    }}
    \end{minipage}
    % \hspace{-0.2cm}
   \begin{minipage}{0.12\linewidth}
     \centering     \includegraphics[width=0.99\linewidth]{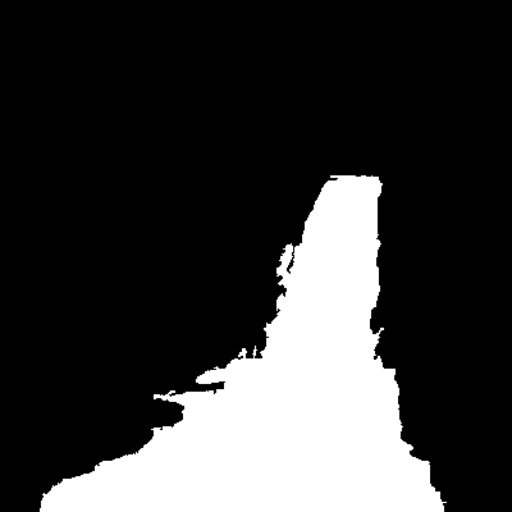}
    \end{minipage}
    \begin{minipage}{0.12\linewidth}
     \centering            \includegraphics[width=0.99\linewidth]{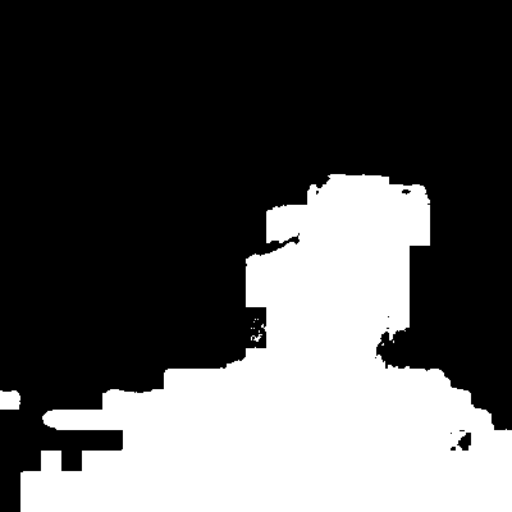}
    \end{minipage}
    \begin{minipage}{0.12\linewidth}
     \centering             \includegraphics[width=0.99\linewidth]{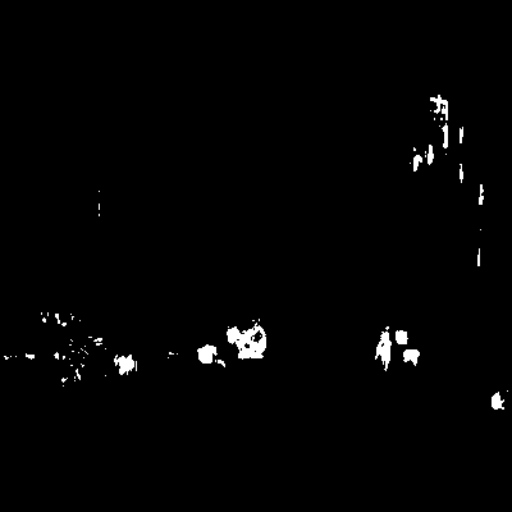}
    \end{minipage}
        \hfill
    \textbf{\vline}
    \hfill    
     \begin{minipage}{0.12\linewidth}
     \centering             
     \includegraphics[width=0.99\linewidth]{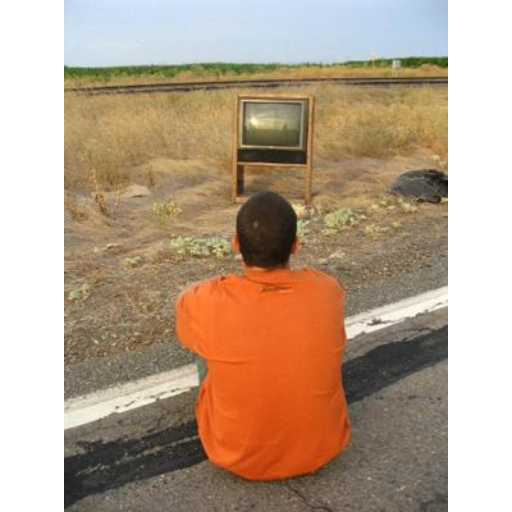}\llap{\raisebox{1.35cm}{%  move next graphics to top right corner
      \includegraphics[height=0.7cm]{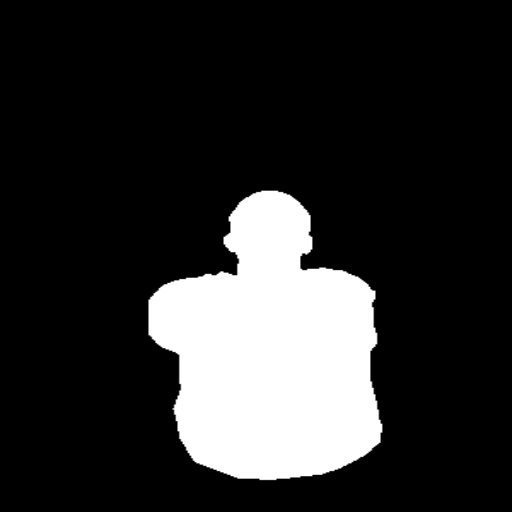}%
    }}
    \end{minipage}
    % \hspace{-0.2cm}
   \begin{minipage}{0.12\linewidth}
     \centering             
     \includegraphics[width=0.99\linewidth]{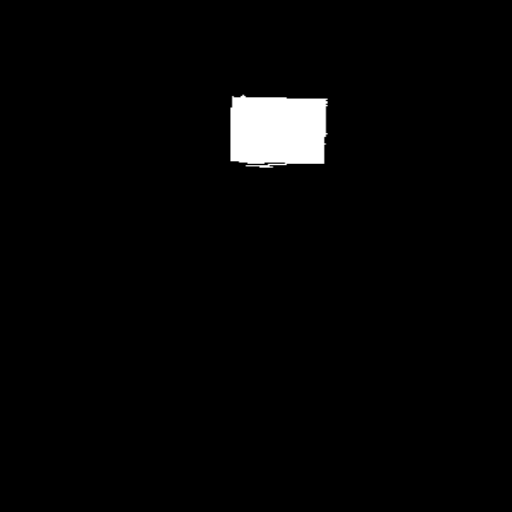}
    \end{minipage}
    \begin{minipage}{0.12\linewidth}
     \centering            \includegraphics[width=0.99\linewidth]{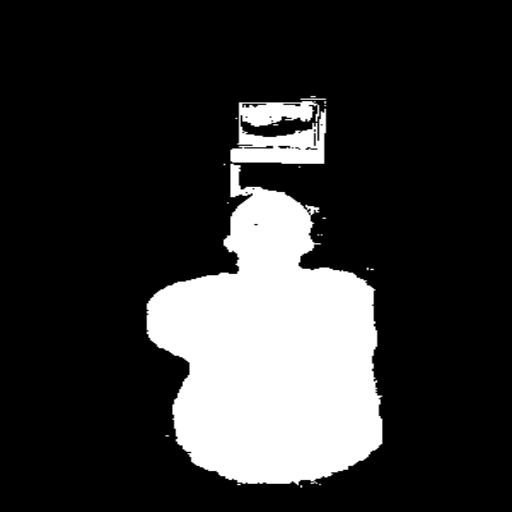}
    \end{minipage}
    \begin{minipage}{0.12\linewidth}
     \centering             \includegraphics[width=0.99\linewidth]{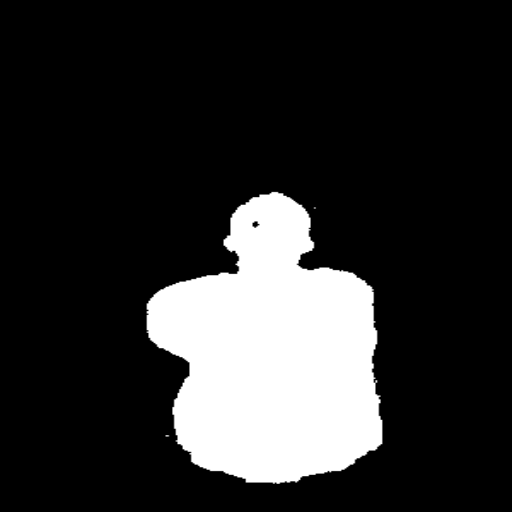}
    \end{minipage}
      \begin{minipage}{0.12\linewidth}
     \centering             
     \includegraphics[width=0.99\linewidth]{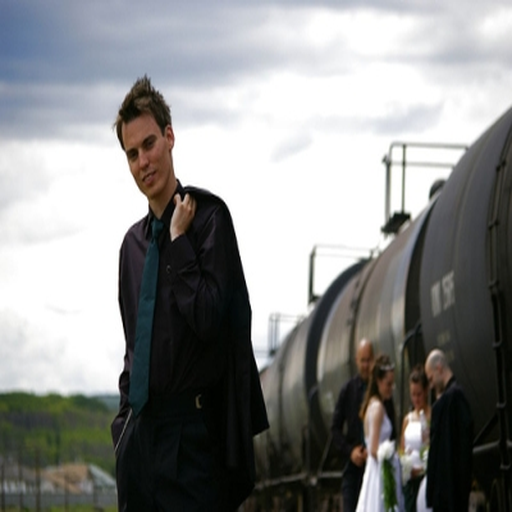}\llap{\raisebox{1.35cm}{%  move next graphics to top right corner
      \includegraphics[height=0.7cm]{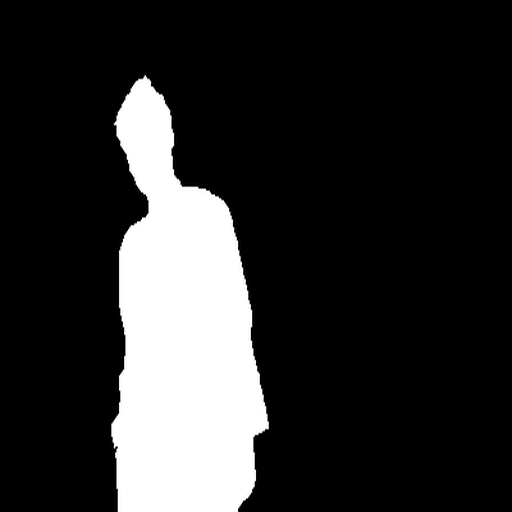}%
    }}
    \end{minipage}
    % \hspace{-0.2cm}
   \begin{minipage}{0.12\linewidth}
     \centering     \includegraphics[width=0.99\linewidth]{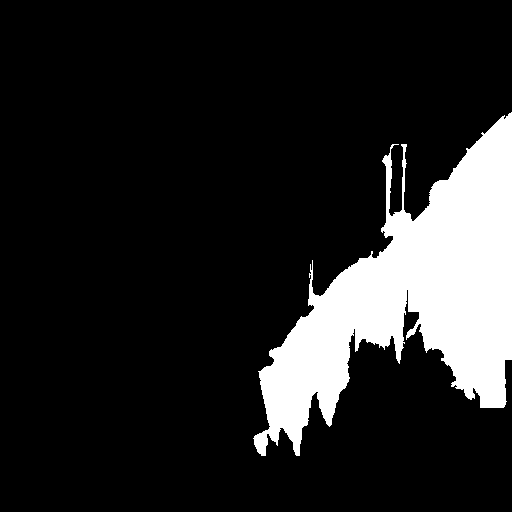}
    \end{minipage}
    \begin{minipage}{0.12\linewidth}
     \centering            \includegraphics[width=0.99\linewidth]{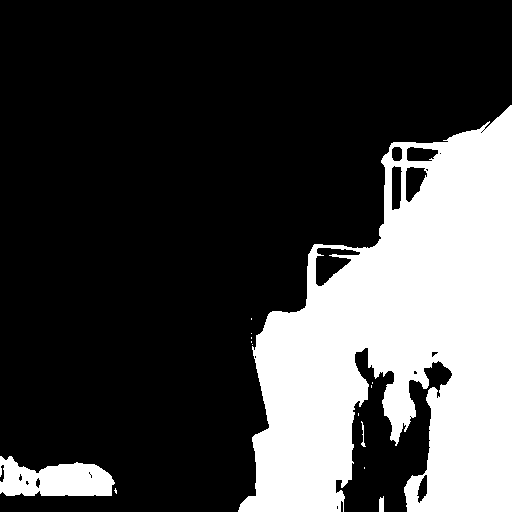}
    \end{minipage}
    \begin{minipage}{0.12\linewidth}
     \centering             \includegraphics[width=0.99\linewidth]{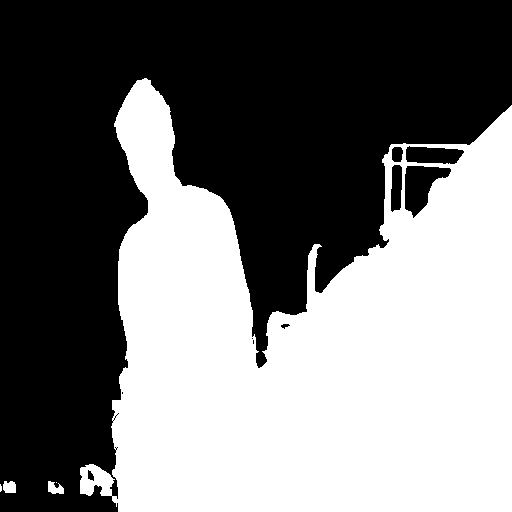}
    \end{minipage}
    \hfill
    \textbf{\vline}
    \hfill    
    \begin{minipage}{0.12\linewidth}
     \centering             
     \includegraphics[width=0.99\linewidth]{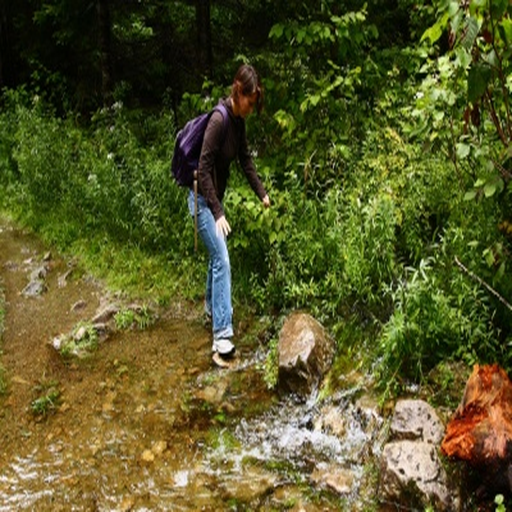}\llap{\raisebox{1.35cm}{%  move next graphics to top right corner
      \includegraphics[height=0.7cm]{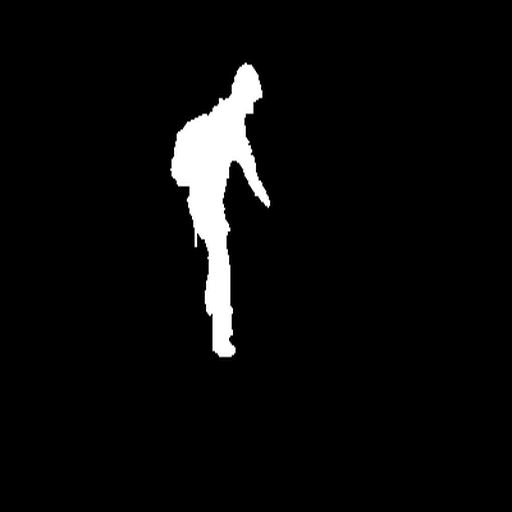}%
    }}
    \end{minipage}
    % \hspace{-0.2cm}
   \begin{minipage}{0.12\linewidth}
     \centering             
     \includegraphics[width=0.99\linewidth]{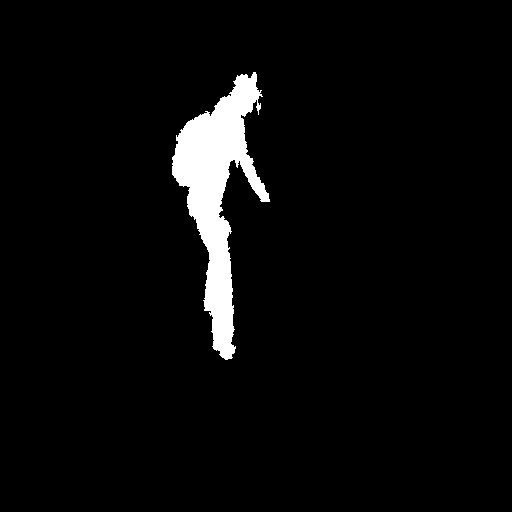}
    \end{minipage}
    \begin{minipage}{0.12\linewidth}
     \centering            \includegraphics[width=0.99\linewidth]{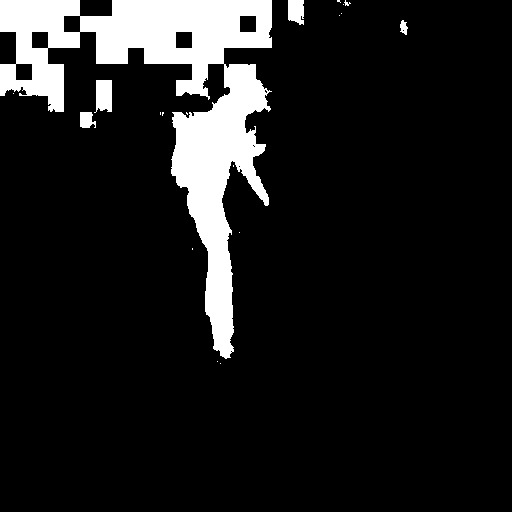}
    \end{minipage}
    \begin{minipage}{0.12\linewidth}
     \centering             \includegraphics[width=0.99\linewidth]{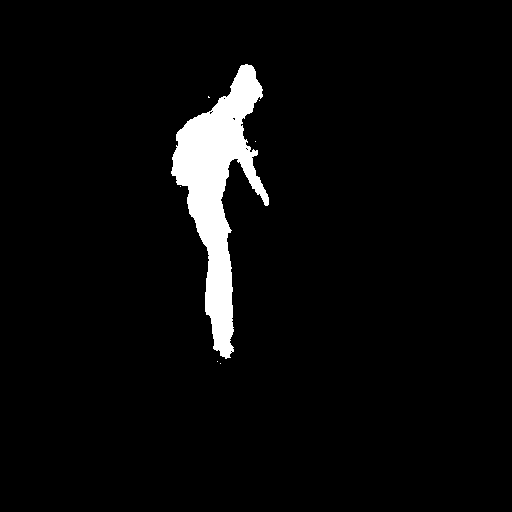}
    \end{minipage}
    \caption{Segmentation Masks obtained with our method and baseline methods (Deep Cut \cite{Aflalo_2023_ICCV} and Deep Spectral Methods (DSM) \cite{melas2022deep}). Ground Truth masks are given on the top-right of each input image.}
    \label{fig:masks}
\end{figure*}
\begin{table*}[!ht]
\centering 
\small
\def\arraystretch{1.18}%
\begin{tabular}{|l|c|c|c|c|c|c|}
\hline
\textbf{Datasets} & \multicolumn{1}{l|}{\textbf{Voynov et al.} \cite{voynov2021object}} & \multicolumn{1}{l|}{\textbf{Melas-Kyriazi et al.}\cite{melas-kyriazi2022finding}} & \multicolumn{1}{l|}{\textbf{TokenCut} \cite{wang2022self}} & \multicolumn{1}{l|}{\textbf{DSM}\cite{melas2022deep}} & \multicolumn{1}{l|}{\textbf{Deep Cut} \cite{Aflalo_2023_ICCV}} & \multicolumn{1}{l|}{\textbf{SimSAM (Ours)}} \\ \hline
 ECSSD\cite{shi2015hierarchical}            & 0.684                              & 0.663                                     & 0.712                                  & 0.733                                 & 0.734                         & \textbf{0.762}                              \\ \hline
DUTS\cite{wang2017learning}             & 0.511                              & 0.528                                     & 0.576                                  & 0.567                                 & 0.560                         & \textbf{0.582}                              \\ \hline
OMRON\cite{yang2013saliency}        & 0.464                              & 0.509                                     & 0.533                                  & 0.514                                 & 0.523                         & \textbf{0.542}                              \\ \hline
CUB\cite{WahCUB_200_2011}              & 0.710                              & 0.664                                     & -                                      & 0.769                                 & \textbf{0.777}                         & 0.770                              \\ \hline
\end{tabular}
\caption{Quantitative Comparison of our proposed method with other baseline methods for object segmentation. It could be observed that the proposed method shows significant improvement in mIoU scores on the following datasets: ECSSD, DUTS, and OMRON, and comparable scores on CUB dataset.}
\label{Table:image_seg}
\end{table*}

\noindent \textbf{Loss Function.} Finally, we describe $\mathcal{L}$ to train the projector of encoder and predictor as follows:
    \begin{equation}
        \mathcal{L} = \frac{1}{2}\big[ \mathcal{D}(f^\alpha_i,sg(\delta^\beta_i)) + \mathcal{D}(sg(\delta^\alpha_i),f^\beta_i)\big]
    \end{equation}
Here, $\texttt{sg}(~)$ denotes the stop-grad operation. $\mathcal{D}(\cdot)$ is the negative cosine similarity function, and it is defined as follows:
    \begin{equation}
       \mathcal{D}(f^\alpha_i,\delta^\beta_i) = -\frac{(f_i^{\alpha})}{\lVert (f_i^{\alpha}) \rVert} \cdot \frac{\delta_i^{\beta}}{\lVert \delta_i^{\beta} \rVert}
    \end{equation}
Similarly, we define $\mathcal{D}(f_i^{\beta},\texttt{sg}(\delta_i^{\alpha}))$. Therefore, the loss function computation is symmetric, and the stop-grad is applied considering both views ($\alpha$ and $\beta$). The stop-grad detaches the weights of the projector layer during training and is useful for obtaining better representations \cite{chen2021exploring}.

% The network architecture is asymmetric by the use of a 'predictor' network.
% \hfill
% \break
\noindent \textbf{Semantic Affinity Matrix.} Using symmetric loss $\mathcal{L}$, predictor ($\pi$) learns to output semantic features that are representative of image patches.
%After training the predictor ($\pi$), we consider the predicted features of view $e$ obtained by projecting it to low dimension space $e$. 
Further, we compute semantic affinity matrix $W_{SA}$ in the inference phase as follows:
    \begin{equation}
        W_{SA} = \big[\pi ( \alpha_1), \cdots, \pi (\alpha_n) \big] \big[\pi (\alpha_1), \cdots, \pi (\alpha_n) \big]^T
     \end{equation}
Here, $W_{SA}$ is the semantic affinity matrix.

% \noindent \textit{\textbf{Inference:}}
% \hfill
% \break
\noindent \textbf{Final Affinity Matrix.} \label{sec:comp_fa} We describe the final affinity matrix $W_{feat}$ that is used further in the spectral methods for segmentation. The final affinity ($W_{feat}$) matrix is computed as the sum of SimSAM~ affinity $W_{SA}$ and vanilla affinity matrix $W_{A}$, which is computed from dense features extracted from DINO-ViT ($\phi$).  $W_{feat}$ is defined as follows:  
\begin{equation}
    W_{feat} = W_{A} + \kappa~ W_{SA} 
\end{equation}
$\kappa$ is a tunable parameter, and we found $\kappa=0.1$ gave best scores (Supplementary Material \footnote{https://github.com/chandagrover/SimSAM/blob/main/kamraSupplemetnary.pdf} in Sec.~5).  \shortname~affinity $W_{SA}$ is computed from the predicted semantic features.
\section{Applications}
\label{sec:applications}
Our proposed semantic feature affinity matrix is used for the spectral method to obtain object and semantic segmentation. Formally, given $W_{feat}$, the eigen decomposition of graph laplacian $L=D^\frac{-1}{2}(D-W_{feat})D^\frac{-1}{2}$ is considered for obtaining multiple eigen segments. $y_0, y_1, ...y_{n-1}$.\\~

\noindent \textbf{Single Object Segmentation.}          
% \label{subsec:sigle_object}
We consider only $y_1$ Fiedler eigenvector to obtain the most prominent objects for segmentation. A pairwise
CRF is applied to increase the resolution of coarse segmentations back to the original image resolution.
% \noindent\textbf{Qualitative Outputs.}

\noindent \textit{Qualitative Results.} Fig.~\ref{fig:masks} shows the prominent visualization of masks obtained with different methods. In all of the outputs, segmentation masks obtained with our method are much closer to the ground truth than Deep Spectral and Deep Cut methods. 
% In the last row, the segmentation masks obtained with our method are equivalent to deep cut \cite{empirical2021iccv} but surpass the DSM \cite{melas2022deep}.  

\noindent \textit{Quantitative Results.} We present mIoU scores on ECSSD \cite{shi2015hierarchical}, DUTS \cite{wang2017learning}, OMRON \cite{yang2013saliency} and CUB \cite{WahCUB_200_2011} datasets for object segmentation tasks in Table~\ref{Table:image_seg}. As we can see, our proposed method outperforms the other baseline methods in mIoU scores for multiple datasets (DUTS, ECSSD, DUTS-OMRON). We observe almost equivalent mIoU scores for the CUB dataset as the best performing Deep Cut method \cite{Aflalo_2023_ICCV}. 

\begin{figure}[!ht]
     \begin{minipage}{0.19\linewidth}
     \centering
        Method
    \end{minipage}
     \begin{minipage}{0.19\linewidth}
     \centering
        Mask
    \end{minipage}
    \begin{minipage}{0.19\linewidth}
     \centering
        EV1
    \end{minipage}
    \begin{minipage}{0.19\linewidth}
     \centering
        EV2
    \end{minipage}
    \begin{minipage}{0.19\linewidth}
     \centering
        EV3
    \end{minipage}
    \begin{minipage}{0.19\linewidth}
     \centering
        DSM \cite{melas2022deep}
    \end{minipage}
     \begin{minipage}{0.19\linewidth}
     \centering             
     \includegraphics[width=0.99\linewidth]{images/Masks/0125_resized_srg.png}
    \end{minipage}
    \begin{minipage}{0.19\linewidth}
     \centering             
     \includegraphics[width=0.99\linewidth]{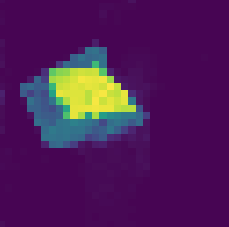}
    \end{minipage}
    \begin{minipage}{0.19\linewidth}
     \centering             
     \includegraphics[width=0.99\linewidth]{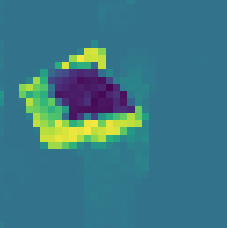}
    \end{minipage}
    \begin{minipage}{0.19\linewidth}
     \centering             
     \includegraphics[width=0.99\linewidth]{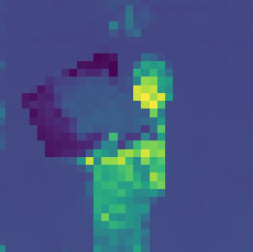}
    \end{minipage}
    \begin{minipage}{0.19\linewidth}
     \centering
        Ours
    \end{minipage}
    \begin{minipage}{0.19\linewidth}
     \centering             
     \includegraphics[width=0.99\linewidth]{images/Masks/0125_resized_ours.png}
    \end{minipage}
    \begin{minipage}{0.19\linewidth}
     \centering             
     \includegraphics[width=0.99\linewidth]{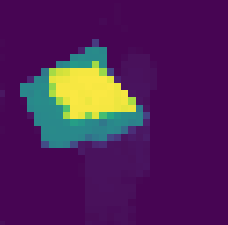}
    \end{minipage}
    \hspace{-0.05cm}
    \begin{minipage}{0.19\linewidth}
     \centering             
     \includegraphics[width=0.99\linewidth]{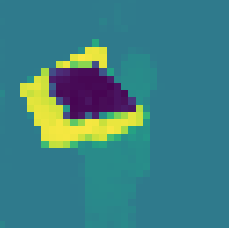}
    \end{minipage}
    \hspace{-0.05cm}
    \begin{minipage}{0.19\linewidth}
     \centering             
     \includegraphics[width=0.99\linewidth]{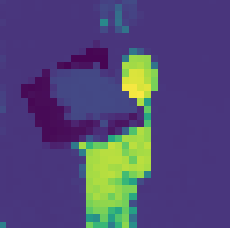}
    \end{minipage}
\caption{\textbf{Eigenvectors (EV)} of an input image (top row left side of Fig.~\ref{fig:masks}). The second column shows the segmentation mask, and the remaining three show the top three eigenvectors.}
    \label{fig:eigen_vector}
\end{figure}

\noindent \textit{Eigenvectors.} Fig.~\ref{fig:eigen_vector} shows the top three eigenvectors for a sample image from Fig.~\ref{fig:masks}. Visibly, the eigenvectors produced by our \shortname~framework (Fig.~\ref{fig:eigen_vector}, 2nd row) provide a better visualization of the complete boy as compared to DSM \cite{melas2022deep}. We considered only the Fiedler vector i.e. EV1 as shown in Fig.~\ref{fig:eigen_vector}, to find coarse object segmentation for computing our semantic affinity matrix, similar to deep spectral methods \cite{melas2022deep}.

\noindent \textbf{Semantic Segmentation.}
We compute 15 smallest eigenvectors for the semantic segmentation task to obtain discrete non-semantic segments for each image. Each image's largest semantic segment is identified as the background region. Following this, a feature vector is computed for each of the non-background segments. The resulting segment features are clustered across the dataset using K-means clustering with K=20 (PASCAL VOC classes). Finally, low-resolution semantic segmentations are produced by associating each segment in the image with its cluster.  

\begin{figure}[!ht]
    \centering
     \begin{minipage}{0.24\linewidth}
     \centering
     \small
        Input
    \end{minipage}
    \centering
     \begin{minipage}{0.24\linewidth}
     \centering
     \small
        DSM \cite{melas2022deep}
    \end{minipage}
     \begin{minipage}{0.24\linewidth}
     \centering
     \small
        SimSAM(ours)
    \end{minipage}
    \begin{minipage}{0.24\linewidth}
     \centering
        \small
        Ground Truth
    \end{minipage}
     \begin{minipage}{0.24\linewidth}
     \centering
        \includegraphics[width=0.99\linewidth]{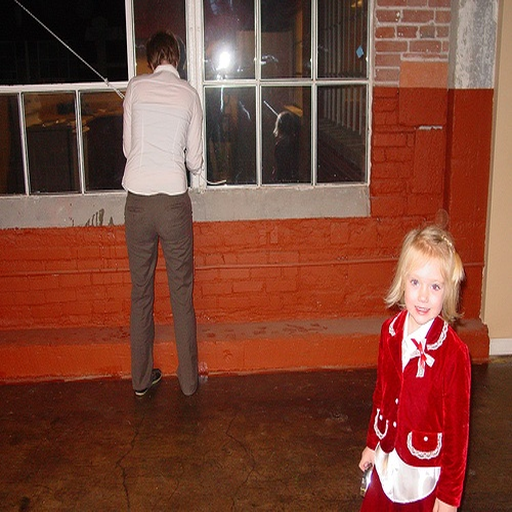}    \end{minipage}
    \begin{minipage}{0.24\linewidth}
     \centering        \includegraphics[width=0.99\linewidth]{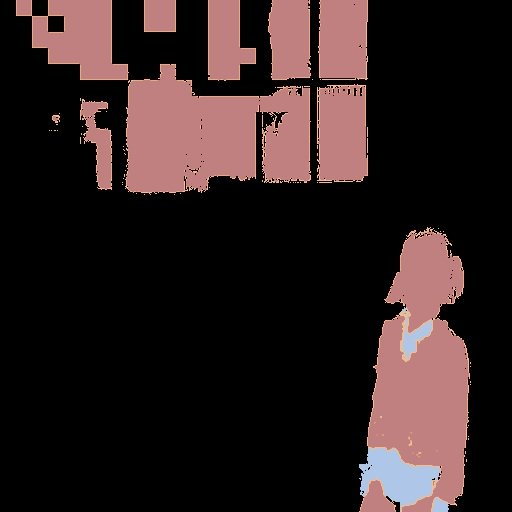}
    \end{minipage}
        \begin{minipage}{0.24\linewidth}
     \centering        \includegraphics[width=0.99\linewidth]{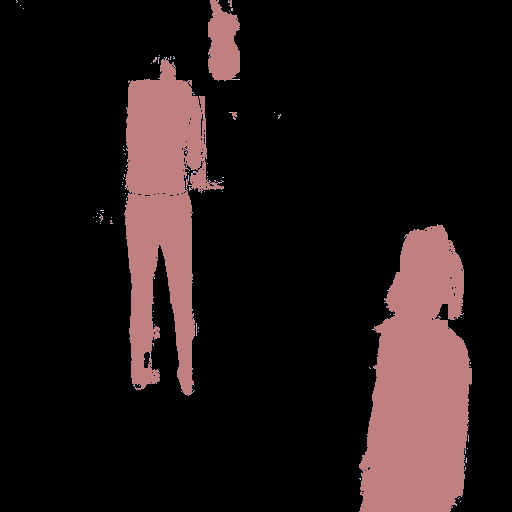}
    \end{minipage}
  \begin{minipage}{0.24\linewidth}
     \centering        \includegraphics[width=0.99\linewidth]{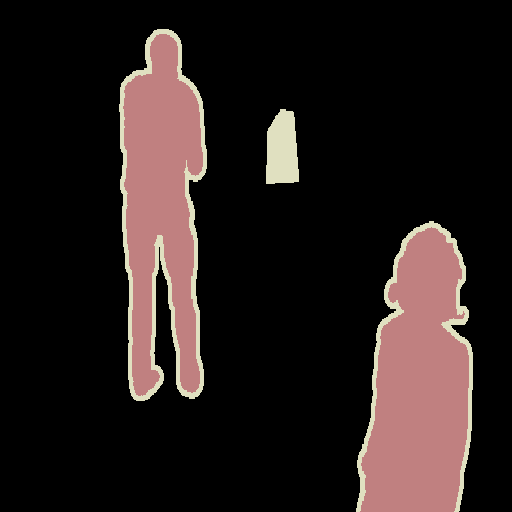}
    \end{minipage}
     \begin{minipage}{0.24\linewidth}
     \centering
        \includegraphics[width=0.99\linewidth]{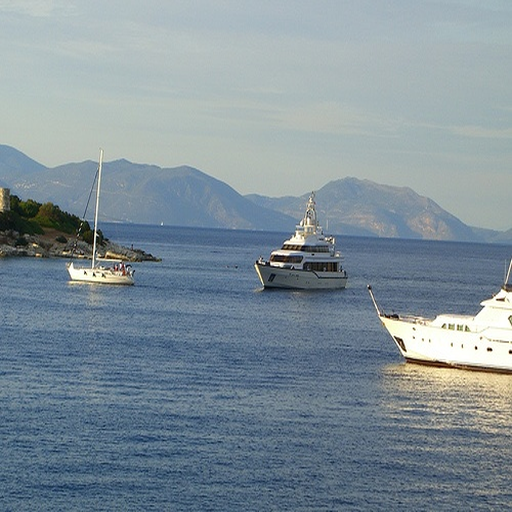}    \end{minipage}
    \begin{minipage}{0.24\linewidth}
     \centering        \includegraphics[width=0.99\linewidth]{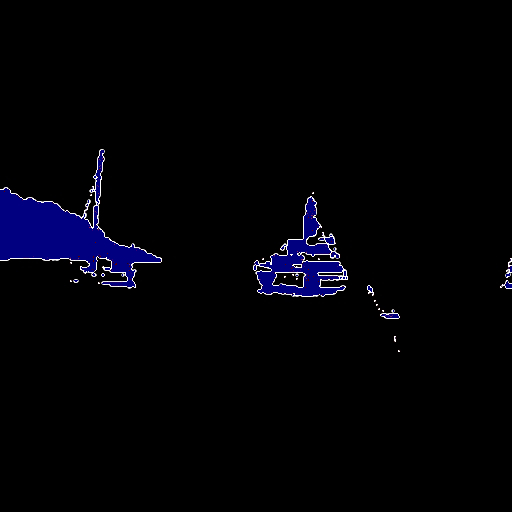}
    \end{minipage}
        \begin{minipage}{0.24\linewidth}
     \centering        \includegraphics[width=0.99\linewidth]{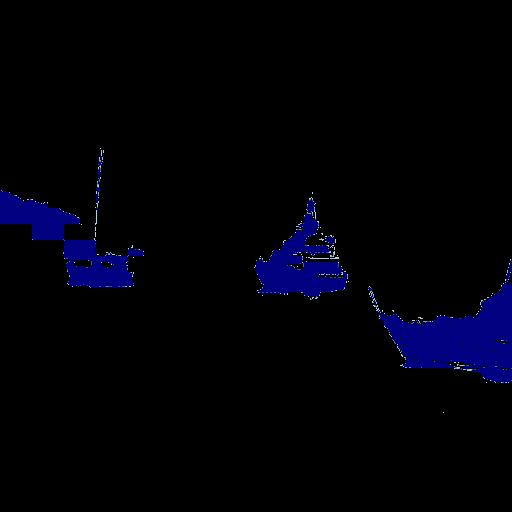}
    \end{minipage}
  \begin{minipage}{0.24\linewidth}
     \centering        \includegraphics[width=0.99\linewidth]{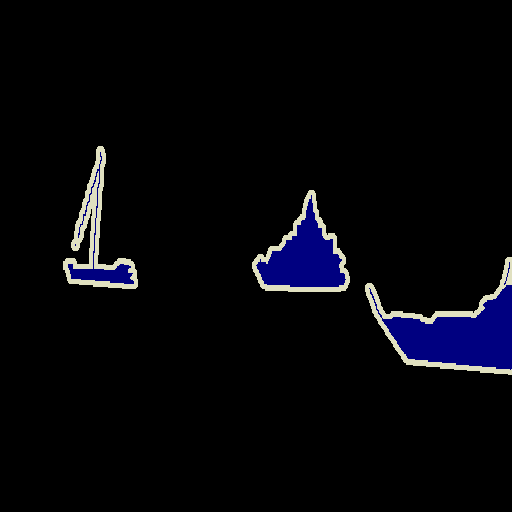}
    \end{minipage}
     \begin{minipage}{0.24\linewidth}
     \centering
        \includegraphics[width=0.99\linewidth]{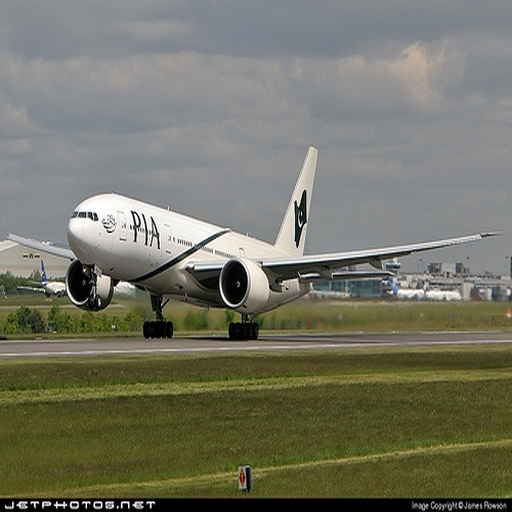}    \end{minipage}
    \begin{minipage}{0.24\linewidth}
     \centering        \includegraphics[width=0.99\linewidth]{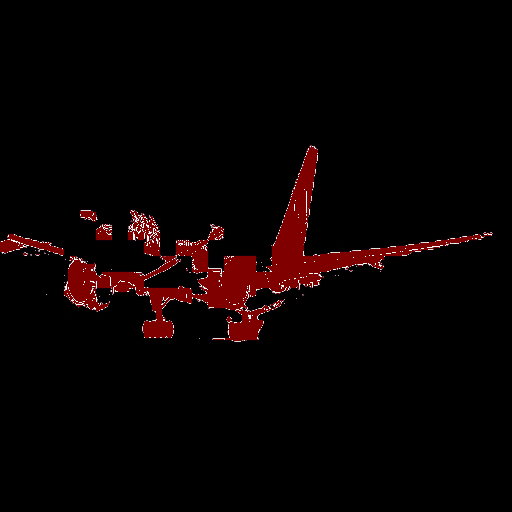}
    \end{minipage}
        \begin{minipage}{0.24\linewidth}
     \centering        \includegraphics[width=0.99\linewidth]{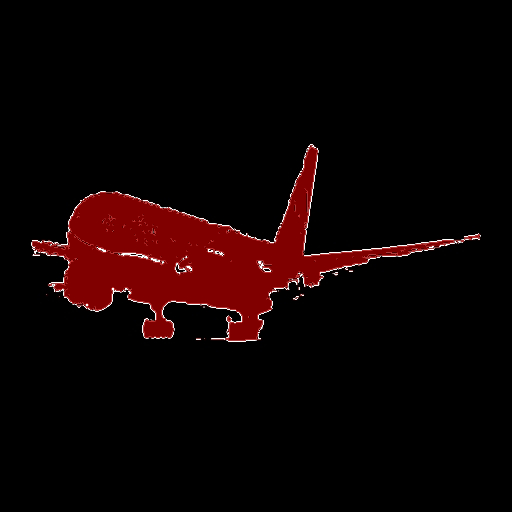}
    \end{minipage}
  \begin{minipage}{0.24\linewidth}
     \centering        \includegraphics[width=0.99\linewidth]{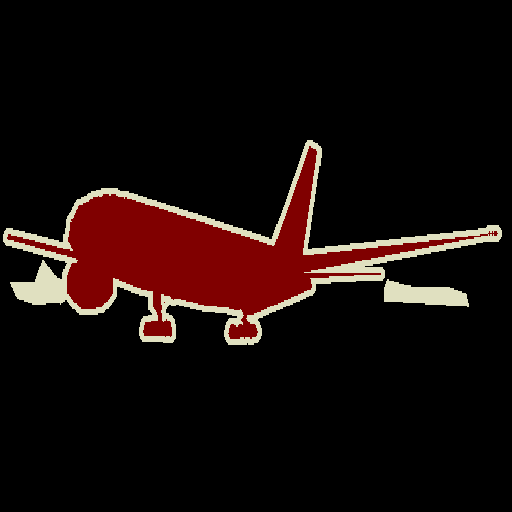}
    \end{minipage}
\caption{Semantic Segmentation Masks obtained with our method and baseline methods (DSM \cite{melas2022deep}).}
\label{fig:sem_segmentation}
 \end{figure}

\noindent \textit{Qualitative Results.} We obtained semantic segmentation masks using our SimSAM framework and compared them against Deep Spectral Methods (DSM) qualitatively, as shown in Fig.~\ref{fig:sem_segmentation}. In all rows, the outputs obtained with our method are very close to ground truth compared to other methods.  

\noindent \textit{Quantitative Results.}
We evaluated our proposed affinity matrix on PASCAL VOC dataset. mIoU scores are reported in Table~\ref{table:semSeg_quant}. The proposed framework outperformed the baseline method by approximately ~3\%.
\begin{table}[!ht]
\centering  \def\arraystretch{1.33}%
\centering
\footnotesize
\begin{tabular}{@{}ccc@{}}
\toprule
\rowcolor[HTML]{FFFFC7} 
MaskContrast & DSM       & \begin{tabular}[c]{@{}c@{}}SimSAM\\ (ours)\end{tabular} \\ \midrule
35                    & 37.2$\pm$3.8 & 40.1$\pm$2.6                                               \\ \bottomrule
\end{tabular}
\caption{Quantitative (mIoU scores) for Semantic Segmentation }
\label{table:semSeg_quant}
\end{table}
\section{Ablation Studies}
\label{sec:Ablation_Study}

\noindent\textbf{Ablation Study-(I).} We performed an ablation study to measure the quality of the semantic affinity matrix computed from SimSAM framework. We randomly selected ten images from the ECSSD \cite{shi2015hierarchical} dataset with their ground-truth masks. We computed each image's affinity matrix with their ground-truth mask, with the mask predicted by DSM \cite{melas2022deep} and our SimSAM framework. Next, we observed the difference between the affinity matrix of ground-truth masks and predicted masks. We found that the average Frobenius norm (F-norm) of SimSAM-based obtained masks (186.99) is closer to ground truth masks than DSM (301.28), as shown in Table~\ref{table:color_ablationstudy}. Improved average accuracy (0.92) and mIoU  (0.68) scores of the selected images signify that the affinity matrix computed from SimSAM framework is semantically more consistent. We provide randomly sampled images for this study and computed scores for each image in the supplementary material \footnote{https://github.com/chandagrover/SimSAM/blob/main/kamraSupplemetnary.pdf}.

\begin{table}[!ht]
\centering  \def\arraystretch{1.15}%
\begin{tabular}{lccc}
\hline
\rowcolor[HTML]{FFFFC7} 
\textbf{Method} & \textbf{\begin{tabular}[c]{@{}c@{}}Frobenius \\ Norm $\downarrow$ \end{tabular}} & \textbf{Accuracy} $\uparrow$ & \textbf{mIoU} $\uparrow$ \\ \hline
DSM \cite{melas2022deep} (w/o SAM)     & 301.28                                                             & 0.86              & 0.46          \\ \hline
Ours (with SAM)   & \textbf{186.99}                                                             & \textbf{0.92}              & \textbf{0.68}          \\ \hline
\end{tabular}
\caption{\textbf{Ablation Study-(I):} Effect of Semantic Affinity on randomly sampled images (similar-looking regions) on ECSSD dataset. The scores reported are averages across ten images.}
\label{table:color_ablationstudy}
\end{table}
for different configurations of the projector and predictor.
\begin{figure*}[!ht]
    \begin{minipage}{0.19\linewidth}
     \centering
     \small
        Non-Linear: Projector and Predictor
    \end{minipage}
    \begin{minipage}{0.19\linewidth}
     \centering
     \small
        Conv1D Projector \\ Non Linear Predictor
    \end{minipage}
    \begin{minipage}{0.19\linewidth}
     \centering
     \small
        Conv1D Projector \\ Conv1D Predictor
    \end{minipage}
    \begin{minipage}{0.19\linewidth}
     \centering
     \small
        PCA Projector \\ Linear Predictor
    \end{minipage}
    \begin{minipage}{0.19\linewidth}
     % \centering
     \small
       Non-Linear Projector \\ Linear Predictor (Ours)
    \end{minipage}
    \begin{minipage}{0.19\linewidth}
     \centering
     \small
        (342.19, 0.85, 0.63)
    \end{minipage}
    \begin{minipage}{0.19\linewidth}
     \centering
     \small
        (222.69, 0.95, 0.85)
    \end{minipage}
    \begin{minipage}{0.19\linewidth}
     \centering
     \small
        (192.18, 0.95, 0.84)
    \end{minipage}
    \begin{minipage}{0.19\linewidth}
     \centering
     \small
        (168.18, 0.97, 0.92)
    \end{minipage}
    \begin{minipage}{0.19\linewidth}
     % \centering
     \small
       \textbf{(157.27, 0.98, 0.95)}
    \end{minipage}
       \begin{minipage}{0.19\linewidth}
     \centering
     \small        
    \end{minipage} 
    \begin{minipage}{0.19\linewidth}
     \centering          
    \includegraphics[width=0.99\linewidth]{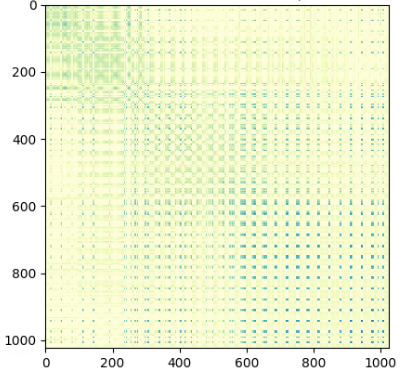}
    \end{minipage}
    % \begin{minipage}{0.32\linewidth}
    %  \centering          
    % \includegraphics[width=0.99\linewidth]{images/ablation_study/con2d+non-linear_W_feat_125.png}
    % \end{minipage}
    \begin{minipage}{0.19\linewidth}
     \centering          
    \includegraphics[width=0.99\linewidth]{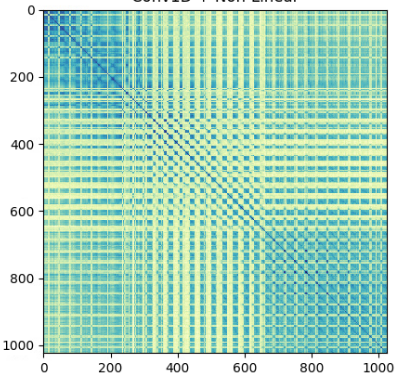}
    \end{minipage}
    \begin{minipage}{0.19\linewidth}
     \centering          
    \includegraphics[width=0.99\linewidth]{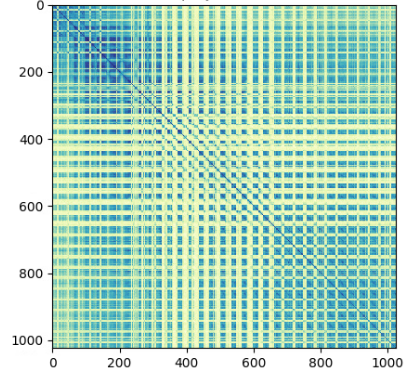}
    \end{minipage}
    \begin{minipage}{0.19\linewidth}
     \centering          
    \includegraphics[width=0.99\linewidth]{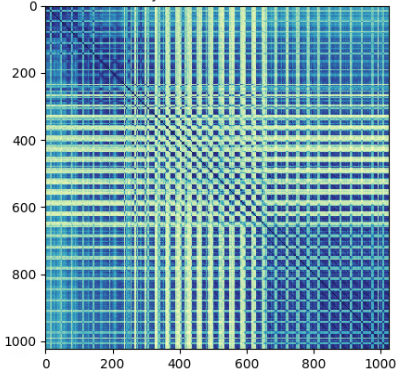}
    \end{minipage}
    %     \begin{minipage}{0.16\linewidth}
    %  \centering          
    % \includegraphics[width=0.99\linewidth]{images/ablation_study/ds_W_feat_125.png}
    % \end{minipage}
        \begin{minipage}{0.19\linewidth}
     \centering          
    \includegraphics[width=0.99\linewidth]{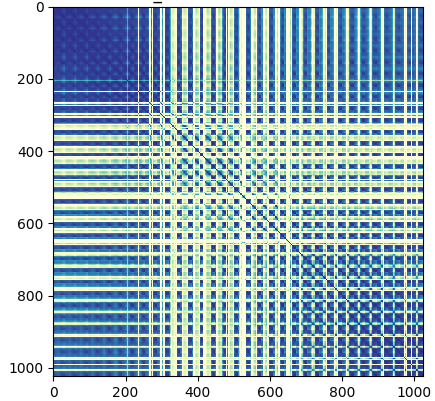}
    \end{minipage}
    \begin{minipage}{0.03\linewidth}
     \centering          
    \includegraphics[width=0.99\linewidth]{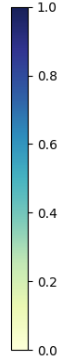}
    \end{minipage}
    \caption{\textbf{Ablation Study-(II):} Feature maps of different configurations of projector and predictor using SimSAM framework. Values reported above each of the feature maps are \textit{(Frobenius Norm, Accuracy, mIoU scores}).}
    \label{Fig:featuremap_ablation}
\end{figure*}
\begin{table*}[!ht]
\centering  
\begin{tabular}{lP{2.5cm}P{2.5cm}P{2.5cm}P{2.5cm}P{2.5cm}}
\hline
\rowcolor[HTML]{FFFFC7} 
\multicolumn{1}{|l|}{\cellcolor[HTML]{FFFFC7}\textbf{Projector}} & \multicolumn{1}{c|}{\cellcolor[HTML]{FFFFC7}Non-linear} & \multicolumn{1}{c|}{\cellcolor[HTML]{FFFFC7}Residual Conv1D} & \multicolumn{1}{c|}{\cellcolor[HTML]{FFFFC7}Residual Conv1D} & \multicolumn{1}{c|}{\cellcolor[HTML]{FFFFC7}PCA}    & \multicolumn{1}{c|}{\cellcolor[HTML]{FFFFC7}\textbf{Non-Linear}} \\ \hline
\rowcolor[HTML]{FFFFC7} 
\multicolumn{1}{|l|}{\cellcolor[HTML]{FFFFC7}\textbf{Predictor}} & \multicolumn{1}{c|}{\cellcolor[HTML]{FFFFC7}Non-linear} & \multicolumn{1}{c|}{\cellcolor[HTML]{FFFFC7}Non-linear}      & \multicolumn{1}{c|}{\cellcolor[HTML]{FFFFC7}Residual Conv1D} & \multicolumn{1}{c|}{\cellcolor[HTML]{FFFFC7}Linear} & \multicolumn{1}{c|}{\cellcolor[HTML]{FFFFC7}\textbf{Linear (Ours)}}       \\ \hline
\textbf{Accuracy}                                                & 0.689                                                   & 0.769                                                        & 0.781                                                        & 0.881                                               & \textbf{0.896 }                                                    \\ \hline
\textbf{mIoU}                                                    & 0.42                                                    & 0.532                                                        & 0.543                                                        & 0.711                                               & \textbf{0.762  }                                                   \\ \hline
\end{tabular}
\caption{\textbf{Ablation Study-(II)}: Different configurations of Siamese Neural Network on ECSSD dataset.}
\label{Table:ablation_study}
\end{table*}

\noindent \textbf{Ablation Study-(II).}
We performed an ablation study of gap analysis (Fig.~\ref{Fig:featuremap_ablation}) of the semantic affinity matrix obtained with different configurations of projector and predictor of Siamese Neural Network on a single image (Fig.~\ref{fig:masks}, top row, first column). For this, we computed the F-norm of the segmentation mask obtained with that configuration corresponding to their ground truth mask. The lowest F-norm score (157.27) of the non-linear projector and linear predictor indicates that this configuration is most similar to the ground truth mask as compared to other configurations. We performed ablation studies for Accuracy and mIoU scores in each configuration, showing that our proposed architecture gives the best scores among other configurations. 

We computed mIoU scores on the entire ECSSD dataset with linear and non-linear configurations, including Residual Conv1D \cite{he2016deep} and PCA \cite{pca_1901}, as shown in Table~\ref{Table:ablation_study}. The best scores are reported with the proposed configuration used in SimSAM framework. 
Visualization of each affinity matrix with respect to the ground truth affinity matrix is present in the supplementary material\footnote{https://github.com/chandagrover/SimSAM/blob/main/kamraSupplemetnary.pdf}.

\noindent\textbf{Ablation Study-(III).} We performed more ablation studies to explore the effect of normalization on the Vanilla affinity matrix $W_{A}= (F_L^K)(F_L^K)^T - \mathcal{M}  \big( (F_L^K)(F_L^K)^T$. Let $W_{A'}=(F_L^K)(F_L^K)^T$ be the matrix without subtracting mean as in $W_{A}$. Table~\ref{Table:ds_ablationStudy} shows that subtracting the mean from the matrix improves the accuracy and mIoU scores. 
\begin{table}[!ht]
\centering  \def\arraystretch{1.18}%
\begin{tabular}{|P{2cm}|P{2cm}|P{2cm}|}
\hline
\rowcolor[HTML]{FFFC9E} 
Affinity Matrix                       & \multicolumn{1}{c|}{\textbf{Accuracy}} & \multicolumn{1}{c|}{\textbf{mIoU}} \\ \hline
$W_{A'}$                                   & \cellcolor[HTML]{FFFFFF}0.859          & \cellcolor[HTML]{FFFFFF}0.678      \\ \hline
\multicolumn{1}{|c|}{$W_{A}$} & \textbf{0.893}                                  & \textbf{0.757}                              \\ \hline
\end{tabular}
\caption{\textbf{Ablation Study-(III)}: Deep Spectral Affinity Matrix on ECSSD dataset. Effect of subtracting mean of feature values from correlation affinity matrix.}
\label{Table:ds_ablationStudy3}
\end{table}

\noindent\textbf{Abalation Study-(IV).}
We considered different for fine-tuning the values of $\kappa$ and found that $\kappa=0.1$ works best for obtaining the best scores on the object segmentation task on ECSSD dataset.
\begin{table}[!ht]
\begin{tabular}{cccccc}
\hline
\rowcolor[HTML]{FFFFC7} 
         & $\kappa$=0.1      & $\kappa$=0.3 & $\kappa$=0.5 & $\kappa$=0.7 & $\kappa$=0.9 \\ \hline
mIoU     & \textbf{0.762} & 0.742     & 0.723     & 0.733     & 0.752     \\ \hline
accuracy & \textbf{0.896} & 0.888     & 0.842     & 0.850     & 0.874     \\ \hline
\end{tabular}
\caption{Abalation Study-(IV): Parametric tuning of $\kappa$ on ECSSD dataset}
\label{table:ablation_study4}
\end{table}
\section{Conclusion}
\label{sec:conclusion}
We propose a framework, \shortname, for image segmentation by incorporating a semantic affinity matrix with the correlation affinity matrix of deep spectral methods utilizing the Simple Siamese framework.
%We used a non-contrastive setting of self-supervised methods, which does not require negative pair examples. 
We conduct various ablation studies on the projector and predictor parts of the Siamese network. We find that a single non-linear layer as a projector and a linear layer as a predictor work best to learn the semantic affinity of pre-trained DINO features. We show that the segmentation mask improves mIoU scores in multiple datasets. In the future, we aim to improve semantic segmentation for identifying more than two objects present in an image.
\bibliographystyle{IEEEbib}
\bibliography{refs}
% \section{Supplementary Material}
\beginsupplement
\twocolumn[{%
\renewcommand\twocolumn[1][]{#1}%
\maketitle
\vspace{-0.3cm}
\begin{minipage}{0.99\linewidth}
\Large
\centering
Supplementary Material
\end{minipage}
\begin{minipage}{0.19\linewidth}
     \centering
     \captionsetup{type=figure}
     \scriptsize
        Non-Linear: Projector and Predictor
    \end{minipage}
    \begin{minipage}{0.19\linewidth}
     \centering
     \scriptsize
        Conv1D Projector \\ Non-Linear Predictor
    \end{minipage}
    \begin{minipage}{0.19\linewidth}
     \centering
     \scriptsize
        Conv1D Projector \\ Conv1D Predictor
    \end{minipage}
    \begin{minipage}{0.19\linewidth}
     \centering
     \scriptsize
        PCA Projector \\ Linear Predictor
    \end{minipage}
    \begin{minipage}{0.21\linewidth}
     \centering
     \scriptsize
       Non-Linear Projector \\ Linear Predictor \textbf{(ours)}
    \end{minipage}
    \begin{minipage}{0.19\linewidth}
     \centering
     \scriptsize
        (342.19, 0.85, 0.63)
    \end{minipage}
    \begin{minipage}{0.19\linewidth}
     \centering
     \scriptsize
        (222.69, 0.95, 0.85)
    \end{minipage}
    \begin{minipage}{0.19\linewidth}
     \centering
     \scriptsize
        (192.18, 0.95, 0.84)
    \end{minipage}
    \begin{minipage}{0.19\linewidth}
     \centering
     \scriptsize
        (168.18, 0.97, 0.92)
    \end{minipage}
    \begin{minipage}{0.19\linewidth}
     \centering
     \scriptsize
       \textbf{(157.27, 0.98, 0.95)}
    \end{minipage}
    % \begin{minipage}{0.19\linewidth}
    %  \centering
    %  \scriptsize        
    % \end{minipage} 
    \begin{minipage}{0.187\linewidth}
     \centering          
    \includegraphics[width=0.99\linewidth]{images/ablation_study/non-linear_W_feat_125_1024.png}
    \end{minipage}
    \begin{minipage}{0.187\linewidth}
     \centering          
    \includegraphics[width=0.99\linewidth]{images/ablation_study/con1d+non-linear_W_feat_125_1024.png}
    \end{minipage}
    \begin{minipage}{0.187\linewidth}
     \centering          
    \includegraphics[width=0.99\linewidth]{images/ablation_study/con1d+con1d_W_feat_125_1024.png}
    \end{minipage}
    \begin{minipage}{0.187\linewidth}
     \centering          
    \includegraphics[width=0.99\linewidth]{images/ablation_study/PCA+LinearPredictor_1024.png}
    \end{minipage}
    \begin{minipage}{0.187\linewidth}
     \centering          
    \includegraphics[width=0.99\linewidth]{images/affinity/W_feat_MatrixOurs_Mask.png}
    \end{minipage}
    \begin{minipage}{0.03\linewidth}
     \centering          
    \includegraphics[width=0.99\linewidth]{images/ablation_study/scale_ablationstudy.png}
    \end{minipage}
    %        \begin{minipage}{0.19\linewidth}
    %  \centering
    %  \scriptsize        
    % \end{minipage} 
    \begin{minipage}{0.187\linewidth}
     \centering          
    \includegraphics[width=0.99\linewidth]{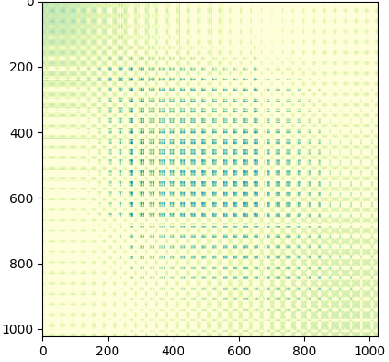}
    \end{minipage}
    \begin{minipage}{0.187\linewidth}
     \centering          
    \includegraphics[width=0.99\linewidth]{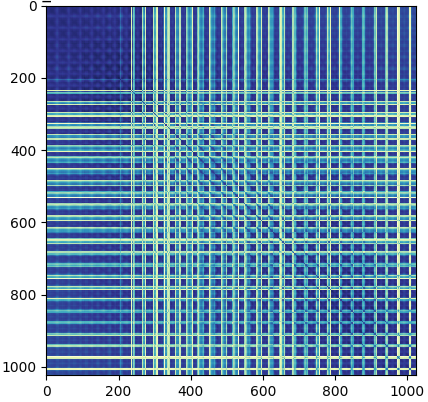}
    \end{minipage}
    \begin{minipage}{0.187\linewidth}
     \centering          
    \includegraphics[width=0.99\linewidth]{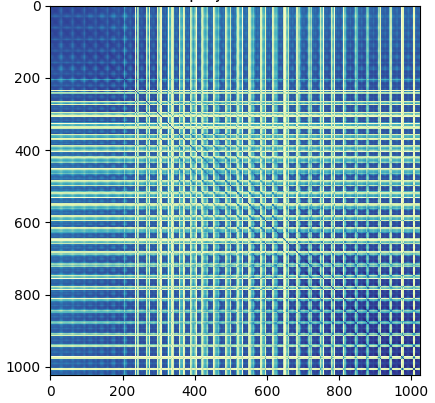}
    \end{minipage}
    \begin{minipage}{0.187\linewidth}
     \centering          
    \includegraphics[width=0.99\linewidth]{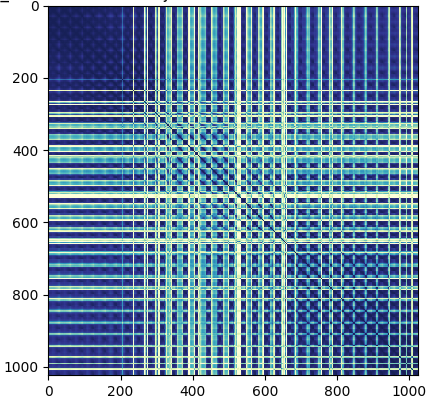}
    \end{minipage}
    \begin{minipage}{0.187\linewidth}
     \centering          
    \includegraphics[width=0.99\linewidth]{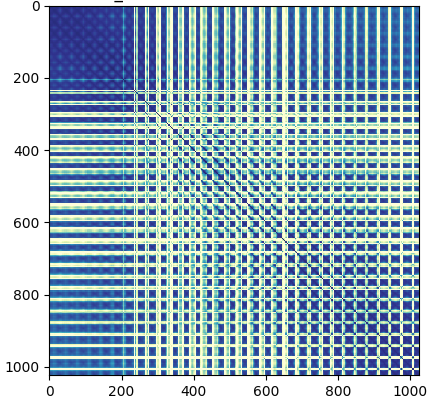}
    \end{minipage}
    \begin{minipage}{0.03\linewidth}
     \centering          
    \includegraphics[width=0.99\linewidth]{images/ablation_study/scale_ablationstudy.png}
    \end{minipage}
    \vspace{-0.3cm}
\captionof{figure}{The top row represents the affinity matrices of the segmentation masks obtained in different configurations of projectors and predictors of a particular image. The bottom row represents the affinity matrices of the ground truth mask. The values reported on top of the figures are \textit{Frobenius Norm, Accuracy and mIOU} scores.}
\label{fig:fig22}
}]
\section{Implementation Details}
Broadly, we present two modules: One is the \textit{\shortname~framework}, which is used for obtaining a semantically consistent, improved semantic matrix (leading to better image segmentation). We trained our Siamese-based SimSAM framework for 10 iterations for each image. 

\noindent \textbf{Projector.}
We experimented with Principal Component Analysis (PCA) as a projector with 64, 128, and 256 components, as shown in Table~\ref{table:pca_component}. 
Among these configurations, 64 components gave the best accuracy and IOU scores for the segmentation mask. We also conducted studies for multiple configurations of the Projector as shown in Fig.~\ref{fig:fig22}. We found that using a non-linear layer as Projector reported the best Frobenius Norm, Accuracy and mIoU scores.\\% When we conducted a study without considering PCA on the Projector and considered only non-linear at the Projector, we found improvement in accuracy and mIoU scores.

\noindent \textbf{Predictor.} We also conducted studies for multiple configurations of predictors on a single image. Fig.~\ref{fig:fig22} shows the affinity matrix obtained in each configuration with respect to its ground truth affinity. As we can see, the linear layer as a predictor reported the best accuracy scores, mIoU and Frobenius norm.
\begin{table}[!ht]
\centering
\begin{tabular}{cccc}
\hline
\rowcolor[HTML]{FFFFC7} 
\textbf{Components} & \textbf{32} & \textbf{64} & \textbf{128} \\ \hline
\textbf{Accuracy}   & 0.77        & \textbf{0.89 }       & 0.88         \\ \hline
\textbf{mIoU}       & 0.54        & \textbf{0.75}        & 0.72         \\ \hline
\end{tabular}
\caption{Number of PCA components. We tested 32, 64 and 128 PCA components. n=64 gave the best scores.}
\label{table:pca_component}
\end{table}

\section{Experimental Results}
\begin{figure*}
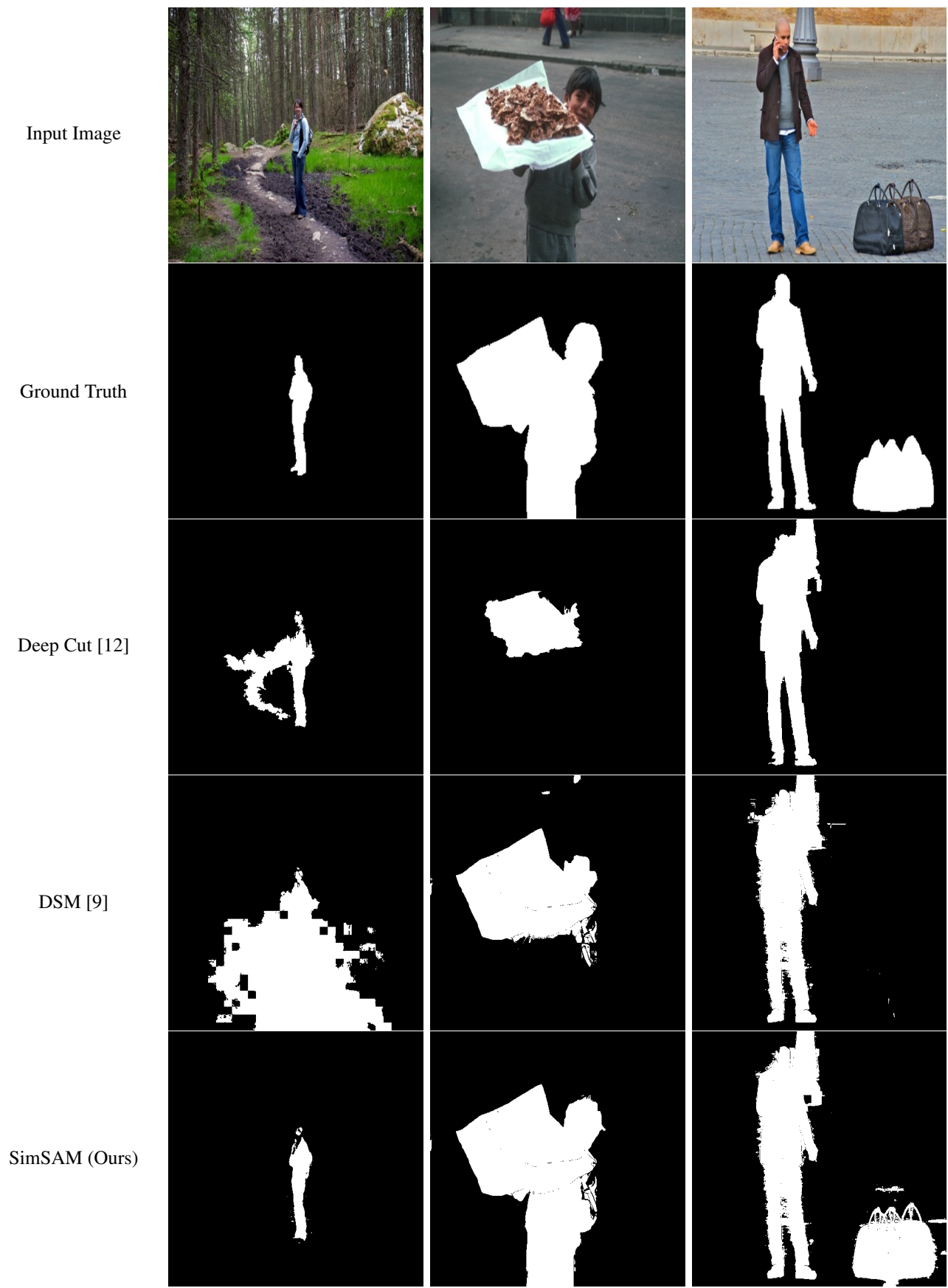

    \begin{minipage}{0.18\linewidth}
     \centering
     Input Image
    \end{minipage}
    \begin{minipage}{0.26\linewidth}
     \centering     \includegraphics[width=0.99\linewidth]{images/Masks/0309_resized.png}
    \end{minipage}
    \begin{minipage}{0.26\linewidth}
     \centering             
     \includegraphics[width=0.99\linewidth]{images/Masks/0125_resized.png}
    \end{minipage}
     \begin{minipage}{0.26\linewidth}
         \centering             
        \includegraphics[width=0.99\linewidth]{images/Masks/0306_resized.png}
    \end{minipage}
   % \begin{minipage}{0.26\linewidth}
   %   \centering             
   %   \includegraphics[width=0.99\linewidth]{images/Masks/0088_resized.png}
   %  \end{minipage}

    \begin{minipage}{0.18\linewidth}
     \centering
    Ground Truth
    \end{minipage}
    \begin{minipage}{0.26\linewidth}
     \centering     \includegraphics[width=0.99\linewidth]{images/Masks/0309_resized_ground.png}
    \end{minipage}
    \begin{minipage}{0.26\linewidth}
     \centering             
     \includegraphics[width=0.99\linewidth]{images/Masks/0125_resized_ground.png}
    \end{minipage}
    \begin{minipage}{0.26\linewidth}
         \centering             
        \includegraphics[width=0.99\linewidth]{images/Masks/0306_resized_ground.png}
    \end{minipage}
   % \begin{minipage}{0.26\linewidth}
   %   \centering             
   %   \includegraphics[width=0.99\linewidth]{images/Masks/0088_resized_ground.png}
   %  \end{minipage}

    \begin{minipage}{0.18\linewidth}
     \centering
    Deep Cut \cite{Aflalo_2023_ICCV}
    \end{minipage}
    \begin{minipage}{0.26\linewidth}
     \centering     \includegraphics[width=0.99\linewidth]{images/Masks/0309_resized_deepcut.png}
    \end{minipage}
    \begin{minipage}{0.26\linewidth}
     \centering             
     \includegraphics[width=0.99\linewidth]{images/Masks/0125_resized_depcut.png}
    \end{minipage}
    \begin{minipage}{0.26\linewidth}
         \centering             
        \includegraphics[width=0.99\linewidth]{images/Masks/0306_resized_deepcut.png}
    \end{minipage}
   % \begin{minipage}{0.26\linewidth}
   %   \centering             
   %   \includegraphics[width=0.99\linewidth]{images/Masks/0088_resized_deepcut.png}
   %  \end{minipage}

    \begin{minipage}{0.18\linewidth}
     \centering
     DSM \cite{melas2022deep}
    \end{minipage}
    \begin{minipage}{0.26\linewidth}
     \centering     \includegraphics[width=0.99\linewidth]{images/Masks/0309_resized_srg.png}
    \end{minipage}
    \begin{minipage}{0.26\linewidth}
     \centering             
     \includegraphics[width=0.99\linewidth]{images/Masks/0125_resized_srg.png}
    \end{minipage}
    \begin{minipage}{0.26\linewidth}
         \centering             
        \includegraphics[width=0.99\linewidth]{images/Masks/0306_resized_srg.png}
    \end{minipage}
   % \begin{minipage}{0.26\linewidth}
   %   \centering             
   %   \includegraphics[width=0.99\linewidth]{images/Masks/0088_resized_srg.png}
   %  \end{minipage}

    \begin{minipage}{0.18\linewidth}
     \centering
     SimSAM (Ours)
    \end{minipage}
    \begin{minipage}{0.26\linewidth}
     \centering     \includegraphics[width=0.99\linewidth]{images/Masks/0309_resized_ours.png}
    \end{minipage}
    \begin{minipage}{0.26\linewidth}
     \centering             
     \includegraphics[width=0.99\linewidth]{images/Masks/0125_resized_ours.png}
    \end{minipage}
    \begin{minipage}{0.26\linewidth}
         \centering             
        \includegraphics[width=0.99\linewidth]{images/Masks/0306_resized_ours.png}
    \end{minipage}
    \caption{Object Segmentation Outputs. SimSAM (ours) is closer to Ground Truth in comparison to baseline methods.}
    \label{fig:fig3}
\end{figure*}

\begin{figure*}
    \begin{minipage}{0.18\linewidth}
     \centering
     Input Image
    \end{minipage}
    \begin{minipage}{0.26\linewidth}
     \centering             
     \includegraphics[width=0.99\linewidth]{images/Masks/sun_aaejodrdtwgvsghs_new.png}
    \end{minipage}
     \begin{minipage}{0.26\linewidth}
         \centering             
        \includegraphics[width=0.99\linewidth]{images/Masks/sun_aawsnvdgguwlkjvq_new.png}
    \end{minipage}
   \begin{minipage}{0.26\linewidth}
     \centering             
     \includegraphics[width=0.99\linewidth]{images/Masks/im103_new.png}
    \end{minipage}

    \begin{minipage}{0.18\linewidth}
     \centering
    Ground Truth
    \end{minipage}
    \begin{minipage}{0.26\linewidth}
     \centering             
     \includegraphics[width=0.99\linewidth]{images/Masks/sun_aaejodrdtwgvsghs_ground_new.png}
    \end{minipage}
     \begin{minipage}{0.26\linewidth}
         \centering             
        \includegraphics[width=0.99\linewidth]{images/Masks/sun_aawsnvdgguwlkjvq_ground_new.png}
    \end{minipage}
   \begin{minipage}{0.26\linewidth}
     \centering             
     \includegraphics[width=0.99\linewidth]{images/Masks/im103_ground_new.png}
    \end{minipage}

    \begin{minipage}{0.18\linewidth}
     \centering
    Deep Cut \cite{Aflalo_2023_ICCV}
    \end{minipage}
    \begin{minipage}{0.26\linewidth}
     \centering             
     \includegraphics[width=0.99\linewidth]{images/Masks/sun_aaejodrdtwgvsghs_deepcut_new.png}
    \end{minipage}
    \begin{minipage}{0.26\linewidth}
         \centering             
        \includegraphics[width=0.99\linewidth]{images/Masks/sun_aawsnvdgguwlkjvq_deepcut_new.png}
    \end{minipage}
   \begin{minipage}{0.26\linewidth}
     \centering             
     \includegraphics[width=0.99\linewidth]{images/Masks/im103_deepcut_new.png}
    \end{minipage}

    \begin{minipage}{0.18\linewidth}
     \centering
     DSM \cite{melas2022deep}
    \end{minipage}
    \begin{minipage}{0.26\linewidth}
     \centering             
     \includegraphics[width=0.99\linewidth]{images/Masks/sun_aaejodrdtwgvsghs_srg_new.png}
    \end{minipage}
    \begin{minipage}{0.26\linewidth}
         \centering             
        \includegraphics[width=0.99\linewidth]{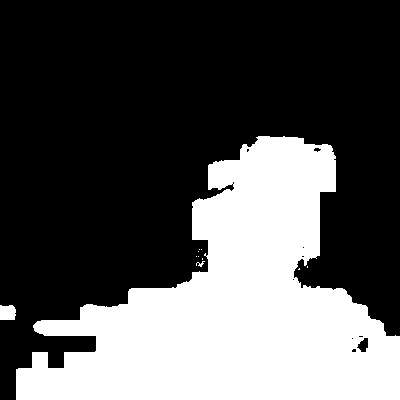}
    \end{minipage}
   \begin{minipage}{0.26\linewidth}
     \centering             
     \includegraphics[width=0.99\linewidth]{images/Masks/im103_srg_new.png}
    \end{minipage}

    \begin{minipage}{0.18\linewidth}
     \centering
     SimSAM (Ours)
    \end{minipage}
    \begin{minipage}{0.26\linewidth}
     \centering             
     \includegraphics[width=0.99\linewidth]{images/Masks/sun_aaejodrdtwgvsghs_ours_new.png}
    \end{minipage}
    \begin{minipage}{0.26\linewidth}
         \centering             
        \includegraphics[width=0.99\linewidth]{images/Masks/sun_aawsnvdgguwlkjvq_ours_new.png}
    \end{minipage}
   \begin{minipage}{0.26\linewidth}
     \centering             
     \includegraphics[width=0.99\linewidth]{images/Masks/im103_ours_new.png}
    \end{minipage}
    \caption{Object Segmentation Outputs. SimSAM (ours) is closer to Ground Truth in comparison to baseline methods.}
    \label{fig:fig4}
\end{figure*}

\begin{figure*}
    \begin{minipage}{0.18\linewidth}
     \centering
    Input Image \cite{Aflalo_2023_ICCV}
    \end{minipage}
   \begin{minipage}{0.26\linewidth}
     \centering             
     \includegraphics[width=0.99\linewidth]{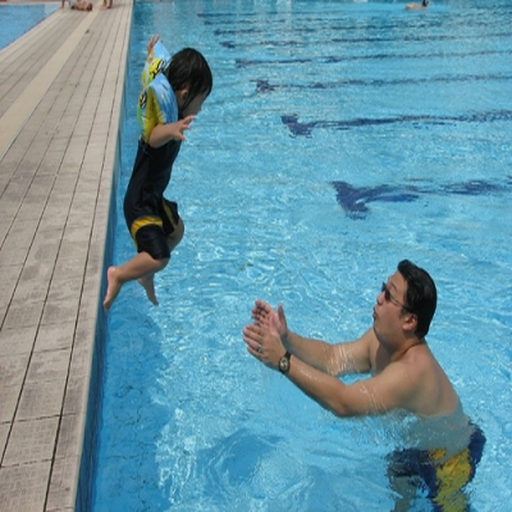}
    \end{minipage}
    \begin{minipage}{0.26\linewidth}
     \centering             
     \includegraphics[width=0.99\linewidth]{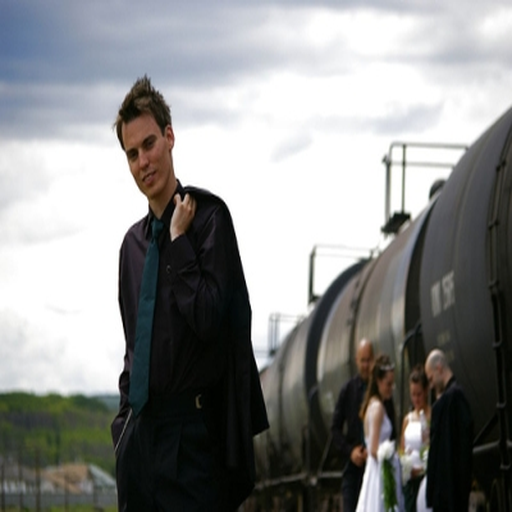}
    \end{minipage} 
    \begin{minipage}{0.26\linewidth}
     \centering             
     \includegraphics[width=0.99\linewidth]{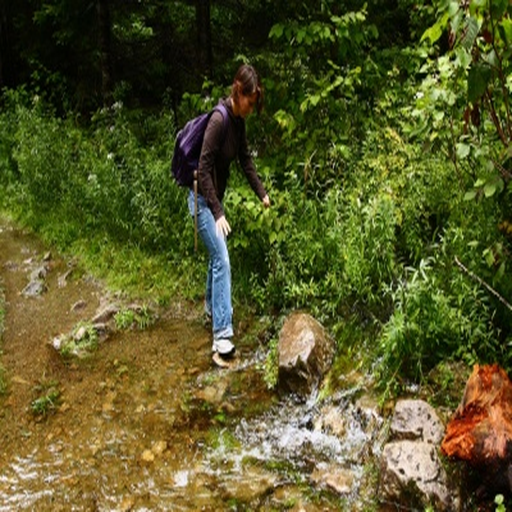}
    \end{minipage} 

    \begin{minipage}{0.18\linewidth}
     \centering
    Ground Truth \cite{Aflalo_2023_ICCV}
    \end{minipage}
   \begin{minipage}{0.26\linewidth}
     \centering             
     \includegraphics[width=0.99\linewidth]{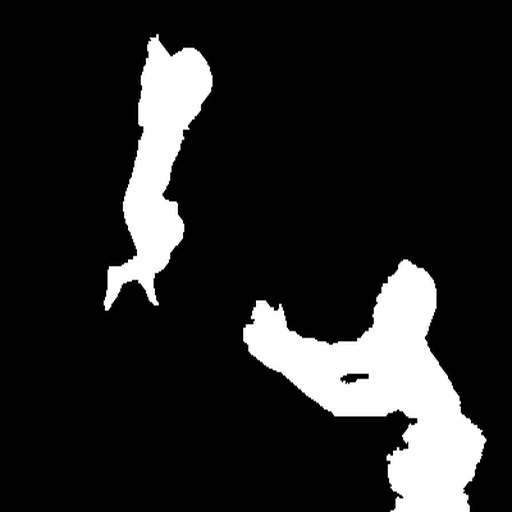}
    \end{minipage}
    \begin{minipage}{0.26\linewidth}
     \centering             
     \includegraphics[width=0.99\linewidth]{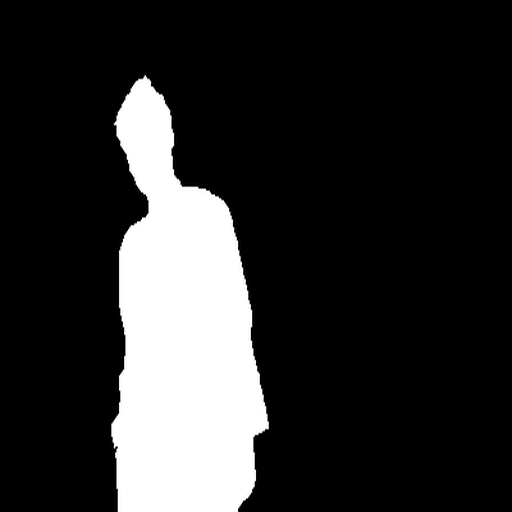}
    \end{minipage} 
    \begin{minipage}{0.26\linewidth}
     \centering             
     \includegraphics[width=0.99\linewidth]{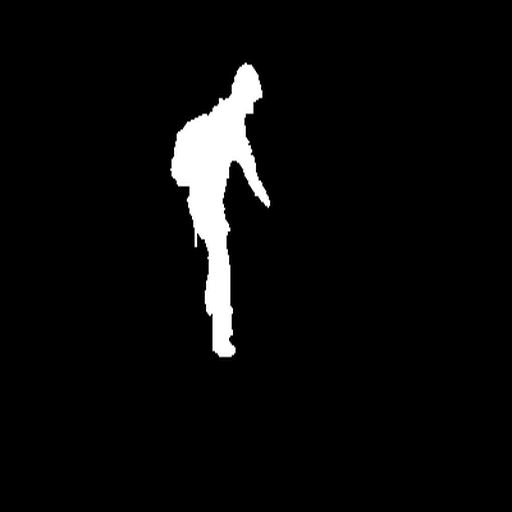}
    \end{minipage} 

    \begin{minipage}{0.18\linewidth}
     \centering
    Deep Cut \cite{Aflalo_2023_ICCV}
    \end{minipage}
   \begin{minipage}{0.26\linewidth}
     \centering             
     \includegraphics[width=0.99\linewidth]{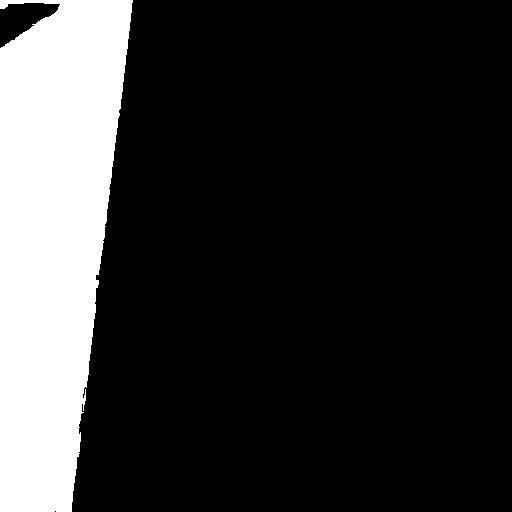}
    \end{minipage}
    \begin{minipage}{0.26\linewidth}
     \centering             
     \includegraphics[width=0.99\linewidth]{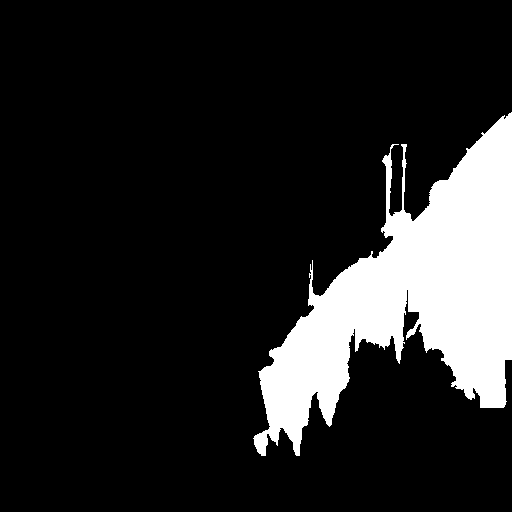}
    \end{minipage} 
    \begin{minipage}{0.26\linewidth}
     \centering             
     \includegraphics[width=0.99\linewidth]{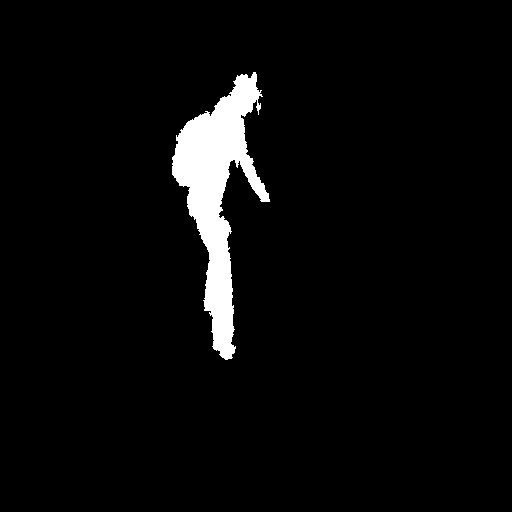}
    \end{minipage} 

    \begin{minipage}{0.18\linewidth}
     \centering
    DSM \cite{melas2022deep}
    \end{minipage}
   \begin{minipage}{0.26\linewidth}
     \centering             
     \includegraphics[width=0.99\linewidth]{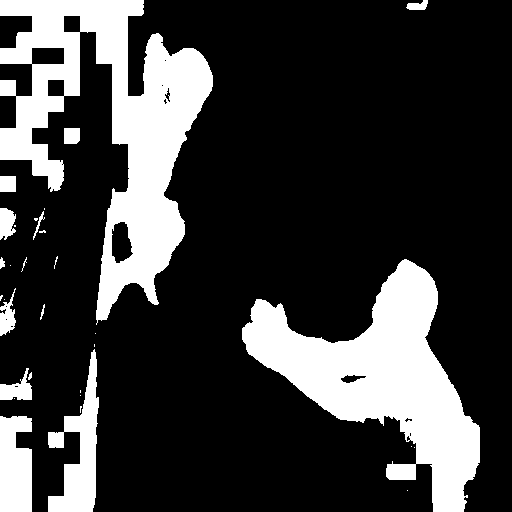}
    \end{minipage}
    \begin{minipage}{0.26\linewidth}
     \centering             
     \includegraphics[width=0.99\linewidth]{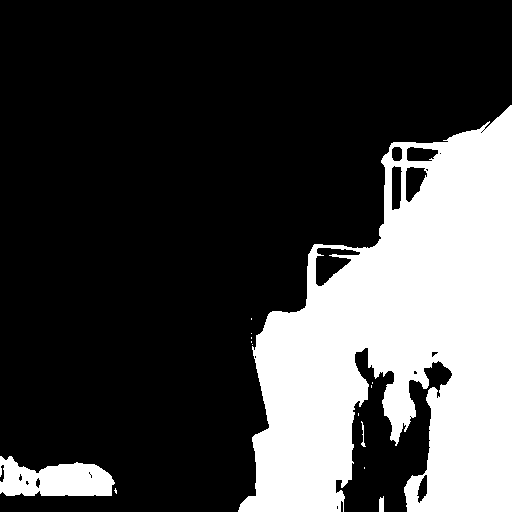}
    \end{minipage} 
    \begin{minipage}{0.26\linewidth}
     \centering             
     \includegraphics[width=0.99\linewidth]{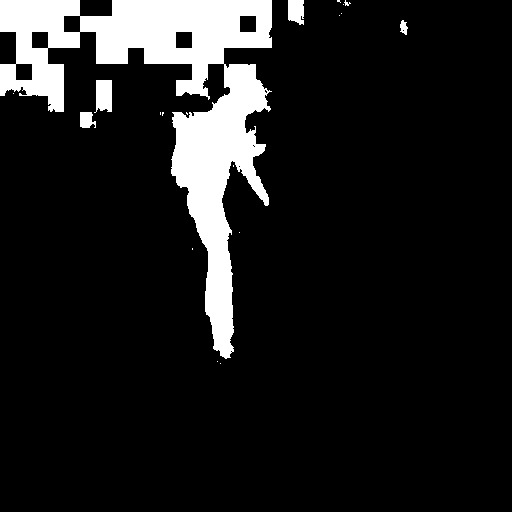}
    \end{minipage} 

    \begin{minipage}{0.18\linewidth}
     \centering
    SimSAM (Ours)
    \end{minipage}
   \begin{minipage}{0.26\linewidth}
     \centering             
     \includegraphics[width=0.99\linewidth]{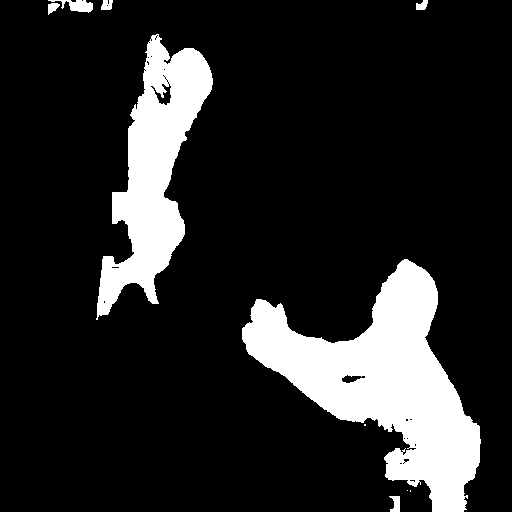}
    \end{minipage}
    \begin{minipage}{0.26\linewidth}
     \centering             
     \includegraphics[width=0.99\linewidth]{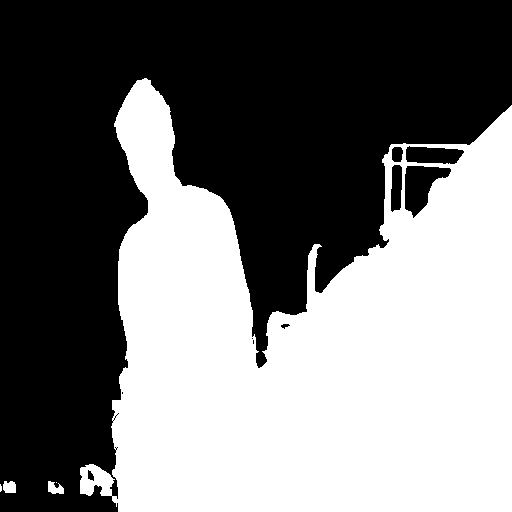}
    \end{minipage} 
    \begin{minipage}{0.26\linewidth}
     \centering             
     \includegraphics[width=0.99\linewidth]{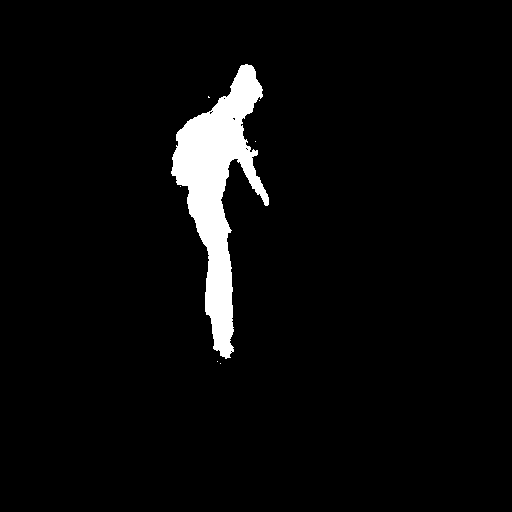}
    \end{minipage} 
    \caption{Object Segmentation Outputs. SimSAM (ours) is closer to Ground Truth in comparison to baseline methods.}
    \label{fig:fig5}   
\end{figure*}

% Table~\ref{table:table_githubcode} represents the code repositories used to produce the outputs for image segmentation (single object segmentation and semantic segmentation). 
% In the main manuscript, we presented the results of three major applications: single-object segmentation, text-based image style transfer, and semantic segmentation.
% We also conducted various experiments for the methodology of our work. 
% We presented small-size outputs in the main manuscript. Here, we present extended results as below:\\ 
\noindent \textbf{Object Segmentation masks Outputs.} Fig.~\ref{fig:fig3}-\ref{fig:fig5} are extended outputs of Fig.~3 of the main manuscript. As shown, the predicted mask obtained with SimSAM (ours) is closest to the ground truth mask as compared to DSM \cite{melas2022deep} and Deep Cut \cite{Aflalo_2023_ICCV}. We reproduced the segmentation masks of baseline methods with the GitHub codes available in DSM \cite{MelasLuke} and DeepCut \cite{AmitAflalo}.

\begin{figure*}
    \begin{minipage}{0.10\linewidth}
     \centering
        Input Image
    \end{minipage}
     \begin{minipage}{0.28\linewidth}
     \centering
        \includegraphics[width=0.99\linewidth]{images/semantic_segmentation/2007_000129.png}
    \end{minipage}
     \begin{minipage}{0.28\linewidth}
     \centering
        \includegraphics[width=0.99\linewidth]{images/semantic_segmentation/2007_002954.png}
    \end{minipage} 
     \begin{minipage}{0.28\linewidth}
     \centering
        \includegraphics[width=0.99\linewidth]{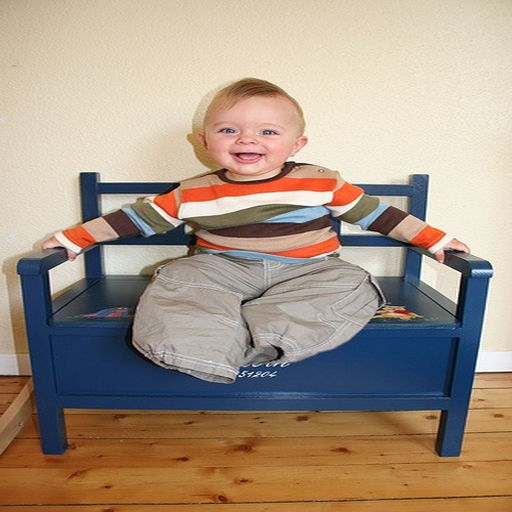}
    \end{minipage} 

 \begin{minipage}{0.10\linewidth}
     \centering
        Ground Truth
    \end{minipage}
     \begin{minipage}{0.28\linewidth}
     \centering
        \includegraphics[width=0.99\linewidth]{images/semantic_segmentation/2007_000129_ground.png}
    \end{minipage}
     \begin{minipage}{0.28\linewidth}
     \centering
        \includegraphics[width=0.99\linewidth]{images/semantic_segmentation/2007_002954_ground.png}
    \end{minipage} 
     \begin{minipage}{0.28\linewidth}
     \centering
        \includegraphics[width=0.99\linewidth]{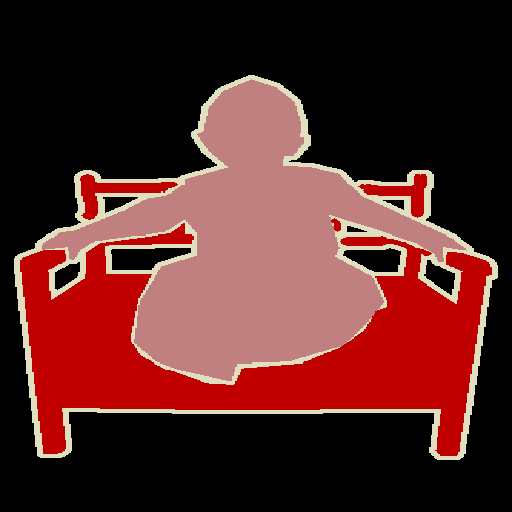}
    \end{minipage} 

 \begin{minipage}{0.10\linewidth}
     \centering
        DSM \cite{melas2022deep}
    \end{minipage}
     \begin{minipage}{0.28\linewidth}
     \centering
        \includegraphics[width=0.99\linewidth]{images/semantic_segmentation/2007_000129_srg_colorm.jpg}
    \end{minipage}
     \begin{minipage}{0.28\linewidth}
     \centering
        \includegraphics[width=0.99\linewidth]{images/semantic_segmentation/2007_002954_srg_colorm.jpg}
    \end{minipage} 
     \begin{minipage}{0.28\linewidth}
     \centering
        \includegraphics[width=0.99\linewidth]{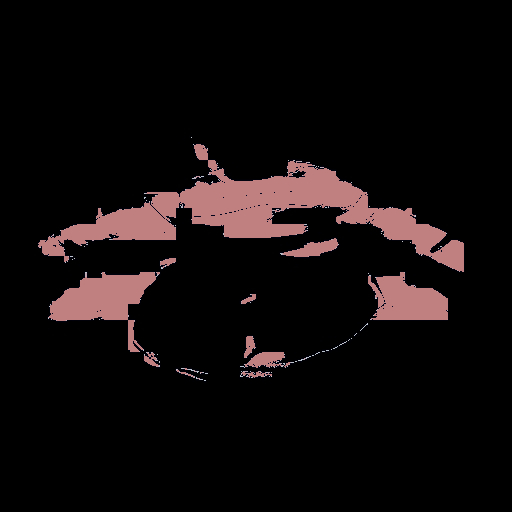}
    \end{minipage} 

 \begin{minipage}{0.10\linewidth}
     \centering
        SimSAM (Ours) 
    \end{minipage}
     \begin{minipage}{0.28\linewidth}
     \centering
        \includegraphics[width=0.99\linewidth]{images/semantic_segmentation/2007_000129_ours_colorm.jpg}
    \end{minipage}
     \begin{minipage}{0.28\linewidth}
     \centering
        \includegraphics[width=0.99\linewidth]{images/semantic_segmentation/2007_002954_srg_colorm.jpg}
    \end{minipage} 
     \begin{minipage}{0.28\linewidth}
     \centering
        \includegraphics[width=0.99\linewidth]{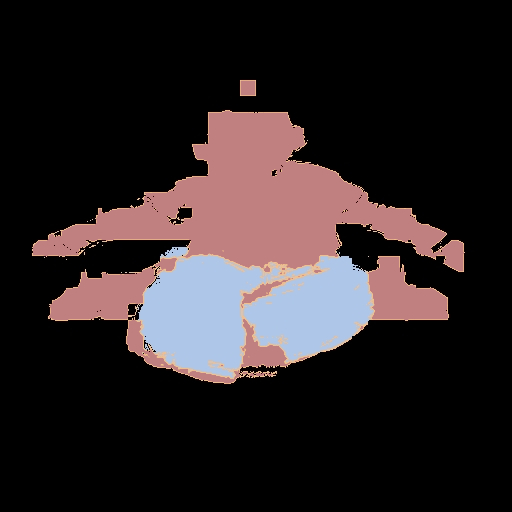}
    \end{minipage}
\caption{Semantic Segmentation Outputs.}
\label{fig:fig_semanticSeg}
\end{figure*}

\begin{figure*}
    \begin{minipage}{0.10\linewidth}
     \centering
        Input Image
    \end{minipage}
     \begin{minipage}{0.28\linewidth}
     \centering
        \includegraphics[width=0.99\linewidth]{images/semantic_segmentation/2007_000241.png}
    \end{minipage}
     \begin{minipage}{0.28\linewidth}
     \centering
        \includegraphics[width=0.99\linewidth]{images/semantic_segmentation/2007_000256.png}
    \end{minipage} 
     \begin{minipage}{0.28\linewidth}
     \centering
        \includegraphics[width=0.99\linewidth]{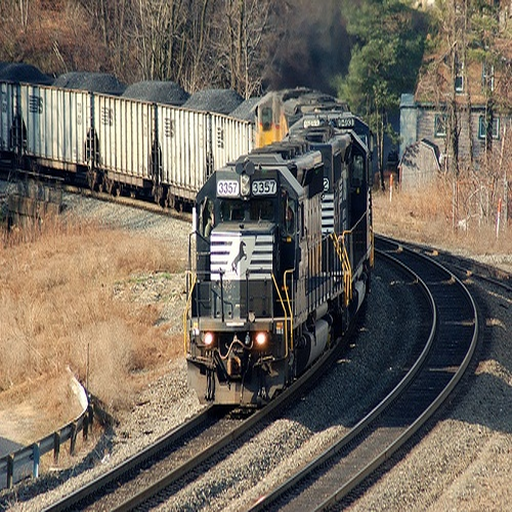}
    \end{minipage} 

 \begin{minipage}{0.10\linewidth}
     \centering
        Ground Truth
    \end{minipage}
     \begin{minipage}{0.28\linewidth}
     \centering
        \includegraphics[width=0.99\linewidth]{images/semantic_segmentation/2007_000241_ground.png}
    \end{minipage}
     \begin{minipage}{0.28\linewidth}
     \centering
        \includegraphics[width=0.99\linewidth]{images/semantic_segmentation/2007_000256_ground.png}
    \end{minipage} 
     \begin{minipage}{0.28\linewidth}
     \centering
        \includegraphics[width=0.99\linewidth]{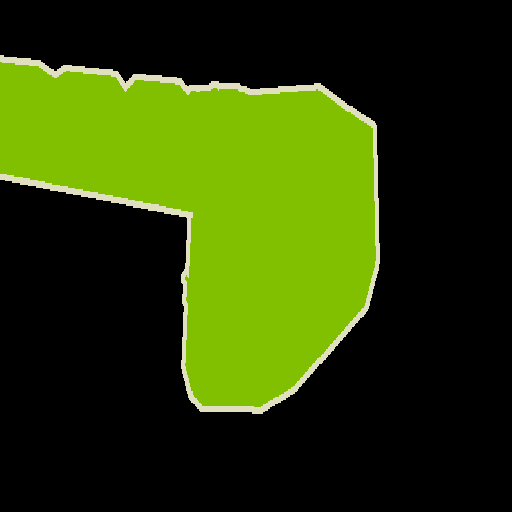}
    \end{minipage} 

 \begin{minipage}{0.10\linewidth}
     \centering
        DSM \cite{melas2022deep}
    \end{minipage}
     \begin{minipage}{0.28\linewidth}
     \centering
        \includegraphics[width=0.99\linewidth]{images/semantic_segmentation/2007_000241_srg_colorm.jpg}
    \end{minipage}
     \begin{minipage}{0.28\linewidth}
     \centering
        \includegraphics[width=0.99\linewidth]{images/semantic_segmentation/2007_000256_srg_colorm.jpg}
    \end{minipage} 
     \begin{minipage}{0.28\linewidth}
     \centering
        \includegraphics[width=0.99\linewidth]{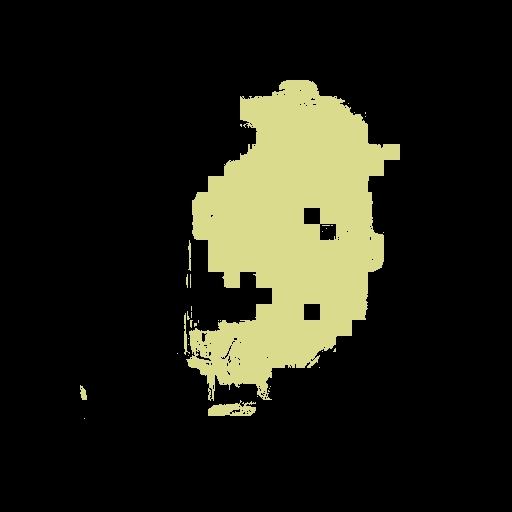}
    \end{minipage} 

 \begin{minipage}{0.10\linewidth}
     \centering
        SimSAM (Ours) 
    \end{minipage}
     \begin{minipage}{0.28\linewidth}
     \centering
        \includegraphics[width=0.99\linewidth]{images/semantic_segmentation/2007_000241_ours_colorm.jpg}
    \end{minipage}
     \begin{minipage}{0.28\linewidth}
     \centering
        \includegraphics[width=0.99\linewidth]{images/semantic_segmentation/2007_000256_ours_colorm.jpg}
    \end{minipage} 
     \begin{minipage}{0.28\linewidth}
     \centering
        \includegraphics[width=0.99\linewidth]{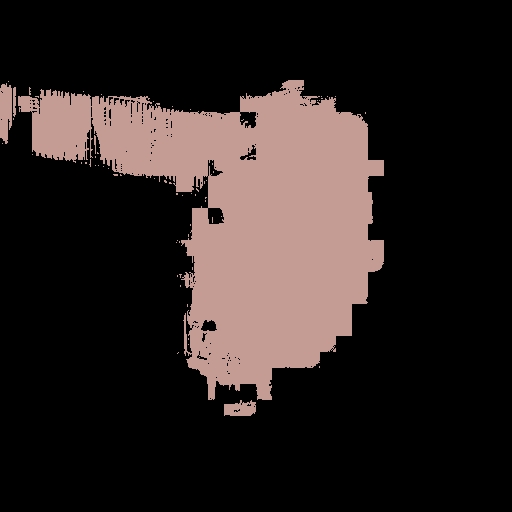}
    \end{minipage}
\caption{Semantic Segmentation Outputs.}
\label{fig:fig_semanticSeg2}
\end{figure*}

\noindent \textbf{Semantic Segmentation Outputs.} Fig.~\ref{fig:fig_semanticSeg} and Fig.~\ref{fig:fig_semanticSeg2} shows the extended results of semantic segmentation masks of Fig.~5 and Fig.~1 of the main manuscript. 

\noindent \textbf{Distinguishable Dense Representations.}  Fig.~\ref{fig:fig20}-\ref{fig:fig21} are detailed scores of individual images whose corresponding average values were reported in Table~3, Ablation Study-(I) of the main manuscript. Semantic Affinity Matrix, Frobenius Norm, mIoU and accuracy scores of 10 randomly sampled images from the ECSSD dataset are presented. The difference between DSM \cite{melas2022deep} and SimSAM is visible in the visualization of affinity matrices with respect to ground truth affinity matrices and with quantitative scores.

\section{Datasets} 
\noindent\textbf{Object Segmentation.} We considered ECSSD \cite{shi2015hierarchical}, DUTS \cite{wang2017learning}, DUTS-OMRON \cite{yang2013saliency} and CUB\cite{WahCUB_200_2011} dataset for training and evaluating the performance of segmentation masks obtained with our method. During training on our SimSAM framework, we considered a batch size of two. During inference, we passed entire image on the trained network to obtain the projected DINO-ViT features for computing a semantically consistent better affinity matrix.  

\noindent \textbf{Semantic Segmentation.}
We performed qualitative and quantitative experiments on the PASCAL VOC dataset for the semantic segmentation task. 
\section{Evaluation Metrics}
\begin{figure*}
    % \begin{center}
    \centering
    \captionsetup{type=figure}
% \begin{center}
     \begin{minipage}{0.21\linewidth}
     \centering
        Input Image
    \end{minipage}
     \begin{minipage}{0.21\linewidth}
     \centering
        Ground Truth
    \end{minipage}
     \begin{minipage}{0.21\linewidth}
     \centering
       DSM \cite{melas2022deep}
    \end{minipage}
     \begin{minipage}{0.21\linewidth}
     \centering
        \shortname~(Ours)
    \end{minipage}
     \begin{minipage}{0.21\linewidth}
     \centering
        \includegraphics[width=0.99\linewidth]{images/Masks/0076_resized.png}
    \end{minipage}
     \begin{minipage}{0.21\linewidth}
     \centering
        \includegraphics[width=0.99\linewidth]{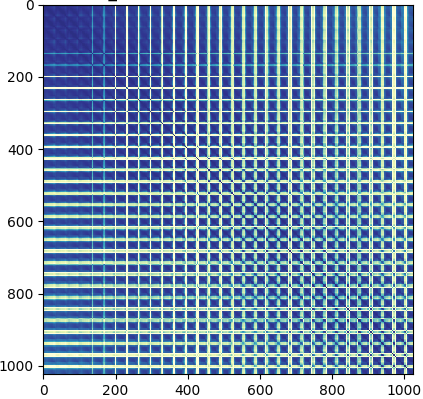}
    \end{minipage}
     \begin{minipage}{0.21\linewidth}
     \centering
        \includegraphics[width=0.99\linewidth]{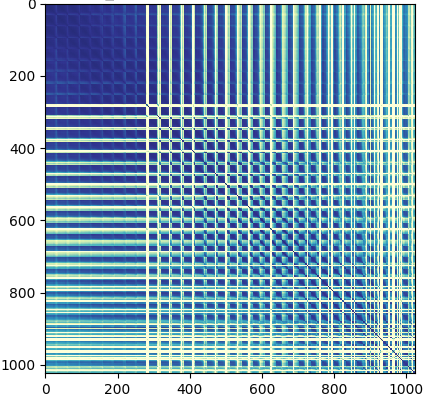}
    \end{minipage}
    \begin{minipage}{0.21\linewidth}
     \centering
        \includegraphics[width=0.99\linewidth]{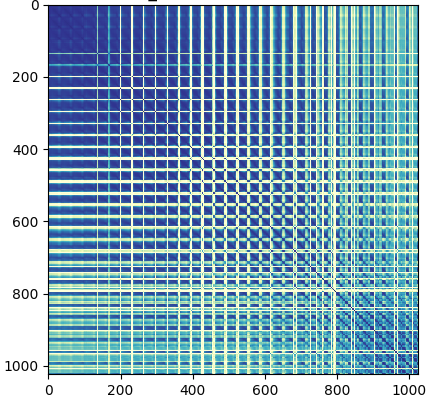}
    \end{minipage} 
    \begin{minipage}{0.04\linewidth}
     \centering          
    \includegraphics[width=0.99\linewidth]{images/ablation_study/scale_ablationstudy.png}
    \end{minipage}
     \begin{minipage}{0.21\linewidth}
     \centering
       ~
    \end{minipage}
    \begin{minipage}{0.21\linewidth}
     \centering
       ~
    \end{minipage}
    \begin{minipage}{0.21\linewidth}
     \centering
     \vspace{-0.12cm}
       (484.70, 0.55, 0.01)
    \end{minipage}
    \begin{minipage}{0.21\linewidth}
     \centering
     \vspace{-0.15cm}
       (231.45, 0.80, 0.47)
    \end{minipage}
     \begin{minipage}{0.21\linewidth}
     \centering
        \includegraphics[width=0.99\linewidth]{images/Masks/0125_resized.png}
    \end{minipage}
     \begin{minipage}{0.21\linewidth}
     \centering
        \includegraphics[width=0.99\linewidth]{images/affinity/W_feat_MatrixGround_Truth_Mask.png}
    \end{minipage}
     \begin{minipage}{0.21\linewidth}
     \centering
        \includegraphics[width=0.99\linewidth]{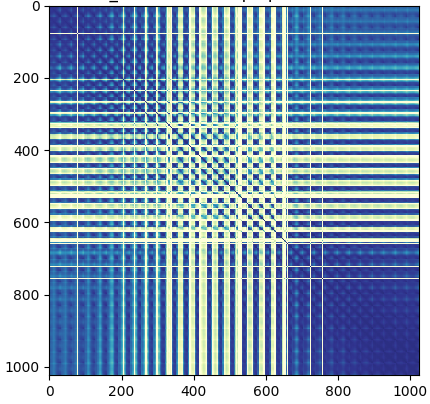}
    \end{minipage}
    \begin{minipage}{0.21\linewidth}
     \centering
        \includegraphics[width=0.99\linewidth]{images/affinity/W_feat_MatrixOurs_Mask.png}
    \end{minipage}  
    \begin{minipage}{0.04\linewidth}
     \centering          
    \includegraphics[width=0.99\linewidth]{images/ablation_study/scale_ablationstudy.png}
    \end{minipage}
     \begin{minipage}{0.21\linewidth}
     \centering
       ~
    \end{minipage}
    \begin{minipage}{0.21\linewidth}
     \centering
       ~
    \end{minipage}
    \begin{minipage}{0.21\linewidth}
     \centering
     \vspace{-0.12cm}
       (312.96, 0.90, 0.62)
    \end{minipage}
    \begin{minipage}{0.21\linewidth}
     \centering
     \vspace{-0.15cm}
       (157.27, 0.93, 0.74)
    \end{minipage}
     \begin{minipage}{0.21\linewidth}
     \centering
        \includegraphics[width=0.99\linewidth]{images/Masks/0309_resized.png}
    \end{minipage}
     \begin{minipage}{0.21\linewidth}
     \centering
        \includegraphics[width=0.99\linewidth]{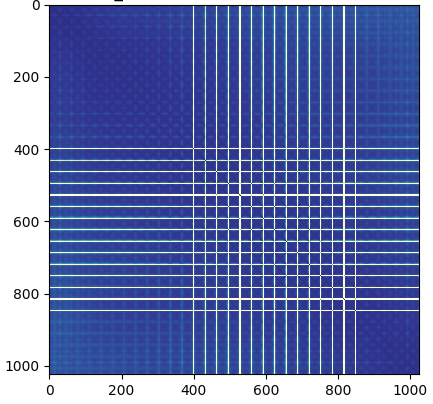}
    \end{minipage}
     \begin{minipage}{0.21\linewidth}
     \centering
        \includegraphics[width=0.99\linewidth]{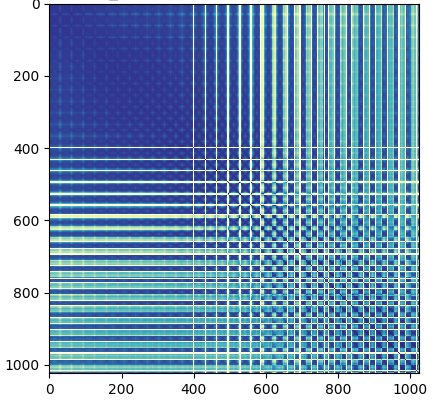}
    \end{minipage}
    \begin{minipage}{0.21\linewidth}
     \centering
        \includegraphics[width=0.99\linewidth]{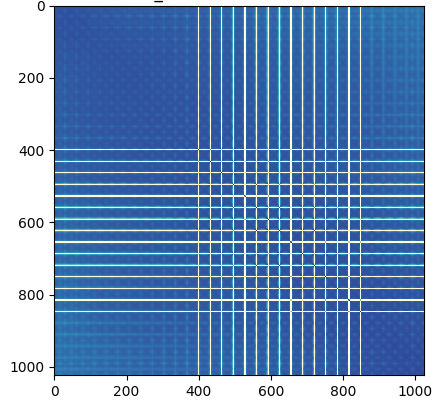}
    \end{minipage}  
    \begin{minipage}{0.04\linewidth}
     \centering          
    \includegraphics[width=0.99\linewidth]{images/ablation_study/scale_ablationstudy.png}
    \end{minipage}
     \begin{minipage}{0.21\linewidth}
     \centering
       ~
    \end{minipage}
    \begin{minipage}{0.21\linewidth}
     \centering
       ~
    \end{minipage}
    \begin{minipage}{0.21\linewidth}
     \centering
     \vspace{-0.12cm}
       (350.81, 0.79, 0.10)
    \end{minipage}
    \begin{minipage}{0.21\linewidth}
     \centering
     \vspace{-0.15cm}
       (132.17, 0.87, 0.35)
    \end{minipage}
     \begin{minipage}{0.21\linewidth}
     \centering
        \includegraphics[width=0.99\linewidth]{images/Masks/0088_resized.png}
    \end{minipage}
     \begin{minipage}{0.21\linewidth}
     \centering
        \includegraphics[width=0.99\linewidth]{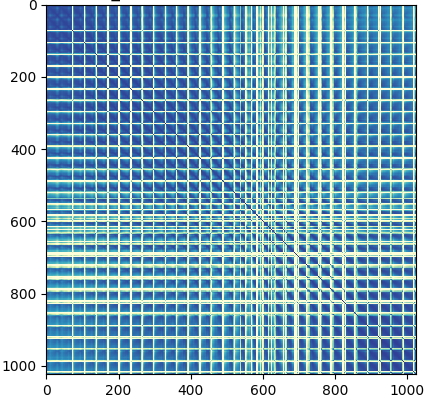}
    \end{minipage}
     \begin{minipage}{0.21\linewidth}
     \centering
        \includegraphics[width=0.99\linewidth]{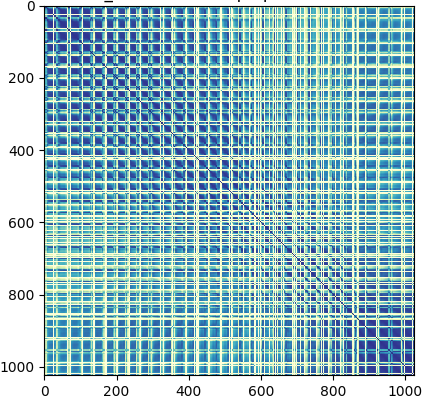}
    \end{minipage}
    \begin{minipage}{0.21\linewidth}
     \centering
        \includegraphics[width=0.99\linewidth]{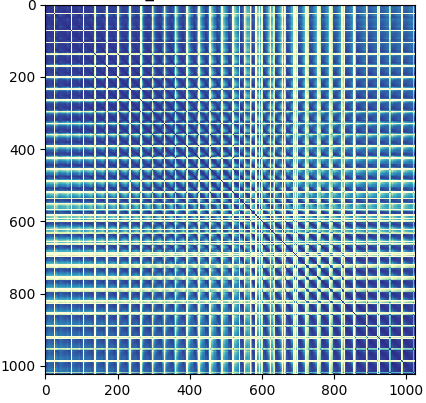}
    \end{minipage}  
    \begin{minipage}{0.04\linewidth}
     \centering          
    \includegraphics[width=0.99\linewidth]{images/ablation_study/scale_ablationstudy.png}
    \end{minipage}
    \begin{minipage}{0.21\linewidth}
     \centering
       ~
    \end{minipage}
    \begin{minipage}{0.21\linewidth}
     \centering
       ~
    \end{minipage}
    \begin{minipage}{0.21\linewidth}
     \centering
     \vspace{-0.12cm}
       (256.06, 0.89, 0.56)
    \end{minipage}
    \begin{minipage}{0.21\linewidth}
     \centering
     \vspace{-0.15cm}
       (114.43, 0.98, 0.86)
    \end{minipage}
     \begin{minipage}{0.21\linewidth}
     \centering
        \includegraphics[width=0.99\linewidth]{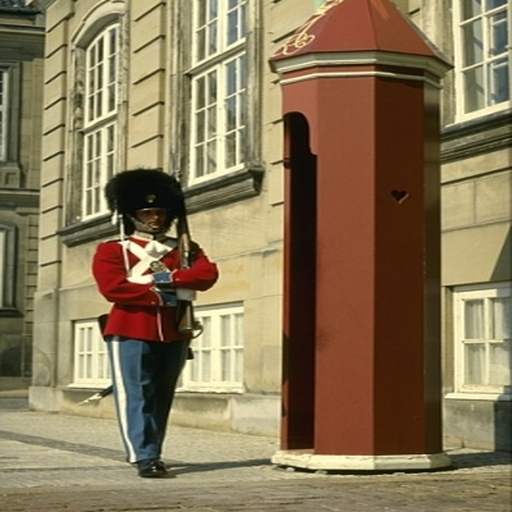}
    \end{minipage}
     \begin{minipage}{0.21\linewidth}
     \centering
        \includegraphics[width=0.99\linewidth]{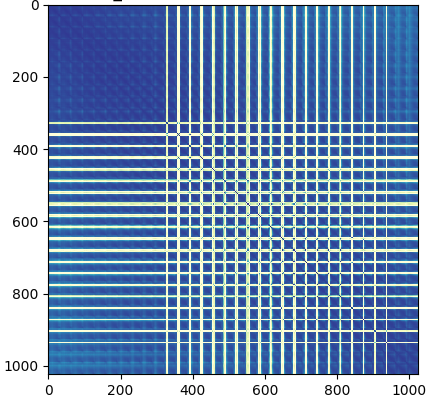}
    \end{minipage}
     \begin{minipage}{0.21\linewidth}
     \centering
        \includegraphics[width=0.99\linewidth]{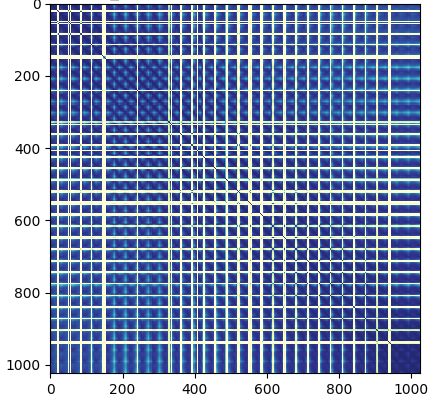}
    \end{minipage}
    \begin{minipage}{0.21\linewidth}
     \centering
        \includegraphics[width=0.99\linewidth]{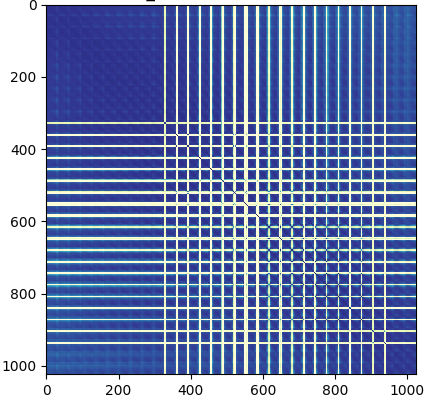}
    \end{minipage} 
    \begin{minipage}{0.04\linewidth}
     \centering          
    \includegraphics[width=0.99\linewidth]{images/ablation_study/scale_ablationstudy.png}
    \end{minipage}
    \begin{minipage}{0.21\linewidth}
     \centering
       ~
    \end{minipage}
    \begin{minipage}{0.21\linewidth}
     \centering
       ~
    \end{minipage}
    \begin{minipage}{0.21\linewidth}
     \centering
     \vspace{-0.2cm}
       (252.19, 0.96, 0.67)
    \end{minipage}
    \begin{minipage}{0.21\linewidth}
     \centering
     \vspace{-0.15cm}
       (124.50, 0.84, 0.33)
    \end{minipage}
    \vspace{-0.3cm}
\caption{Values reported at bottom of DSM \cite{melas2022deep} and SimSAM (ours) method is Frobenius Norm, Accuracy and mIoU scores of randomly sampled image from ECSSD dataset. \textbf{Ablation Study-(I)} of main manuscript.}
\label{fig:fig20}
    % \begin{minipage}{0.95\linewidth}
    %  \centering
    %  \vspace{0.3cm}
    %  Values reported at bottom of each baseline method is Frobenius Norm, Accuracy and mIoU scores of randomly sampled image from ECSSD dataset. Ablation Study I of Paper. 
    \end{figure*}

\begin{figure*}
    \centering
    \captionsetup{type=figure}
     \begin{minipage}{0.21\linewidth}
     \centering
        Input Image
    \end{minipage}
     \begin{minipage}{0.21\linewidth}
     \centering
        Ground Truth
    \end{minipage}
     \begin{minipage}{0.21\linewidth}
     \centering
       DSM \cite{melas2022deep}
    \end{minipage}
     \begin{minipage}{0.21\linewidth}
     \centering
        \shortname~(Ours)
    \end{minipage}
     \begin{minipage}{0.21\linewidth}
     \centering
        \includegraphics[width=0.99\linewidth]{images/Masks/0488_resized.png}
    \end{minipage}
     \begin{minipage}{0.21\linewidth}
     \centering
        \includegraphics[width=0.99\linewidth]{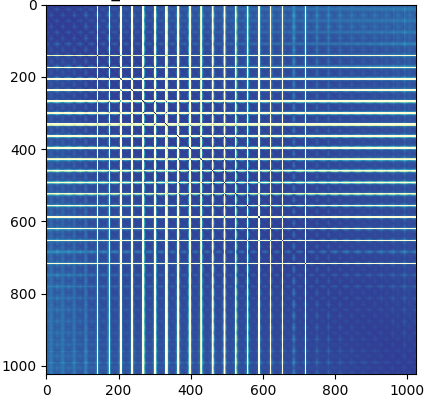}
    \end{minipage}
     \begin{minipage}{0.21\linewidth}
     \centering
    \includegraphics[width=0.99\linewidth]{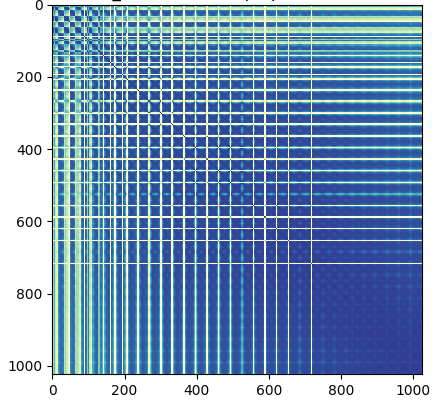}
    \end{minipage}
    \begin{minipage}{0.21\linewidth}
     \centering    \includegraphics[width=0.99\linewidth]{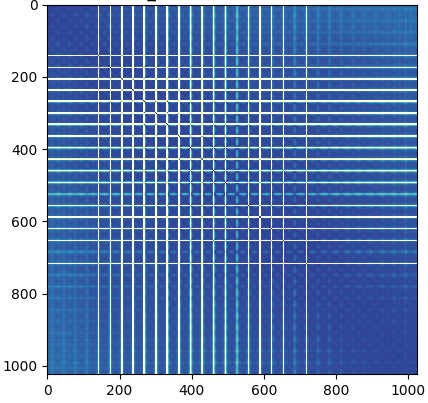}
    \end{minipage}  
    \begin{minipage}{0.04\linewidth}
     \centering         \includegraphics[width=0.99\linewidth]{images/ablation_study/scale_ablationstudy.png}
    \end{minipage} 
    \begin{minipage}{0.21\linewidth}
     \centering
       ~
    \end{minipage}
    \begin{minipage}{0.21\linewidth}
     \centering
       ~
    \end{minipage}
    \begin{minipage}{0.21\linewidth}
     \centering
     \vspace{-0.15cm}
       (215.10, 0.93, 0.36)
    \end{minipage}
    \begin{minipage}{0.21\linewidth}
     \centering
     \vspace{-0.15cm}
       (76.99, 0.99, 0.92)
    \end{minipage}
    \begin{minipage}{0.21\linewidth}
     \centering        \includegraphics[width=0.99\linewidth]{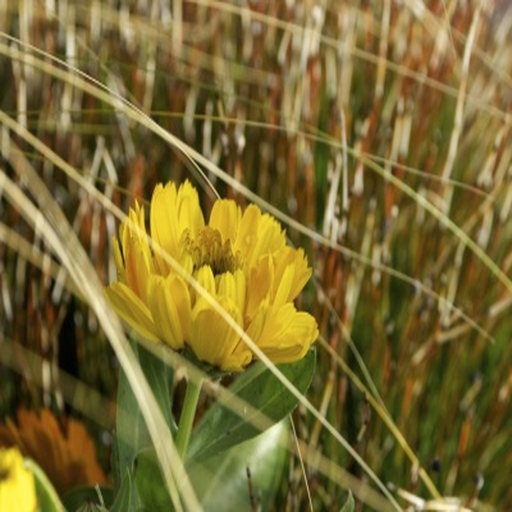}
    \end{minipage}
     \begin{minipage}{0.21\linewidth}
     \centering        \includegraphics[width=0.99\linewidth]{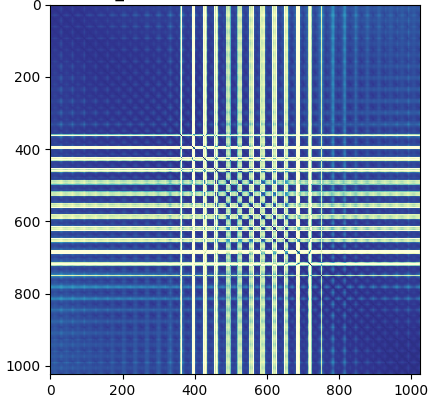}
    \end{minipage}
     \begin{minipage}{0.21\linewidth}
     \centering
        \includegraphics[width=0.99\linewidth]{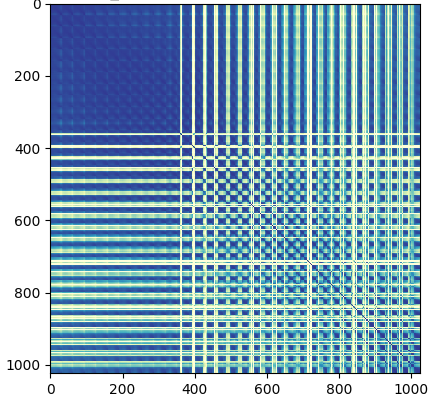}
    \end{minipage}
    \begin{minipage}{0.21\linewidth}
     \centering
        \includegraphics[width=0.99\linewidth]{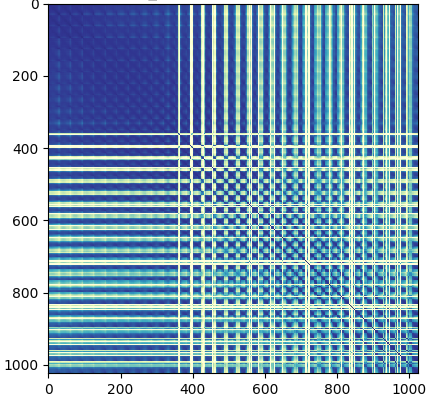}
    \end{minipage} 
    \begin{minipage}{0.04\linewidth}
     \centering          
    \includegraphics[width=0.99\linewidth]{images/ablation_study/scale_ablationstudy.png}
    \end{minipage}
    \begin{minipage}{0.21\linewidth}
     \centering
       ~
    \end{minipage}
    \begin{minipage}{0.21\linewidth}
     \centering
       ~
    \end{minipage}
    \begin{minipage}{0.21\linewidth}
     \centering
     \vspace{-0.15cm}
       (291.89, 0.89, 0.49)
    \end{minipage}
    \begin{minipage}{0.21\linewidth}
     \centering
     \vspace{-0.15cm}
       (275.27, 0.93, 0.69)
    \end{minipage}
     \begin{minipage}{0.21\linewidth}
     \centering
        \includegraphics[width=0.99\linewidth]{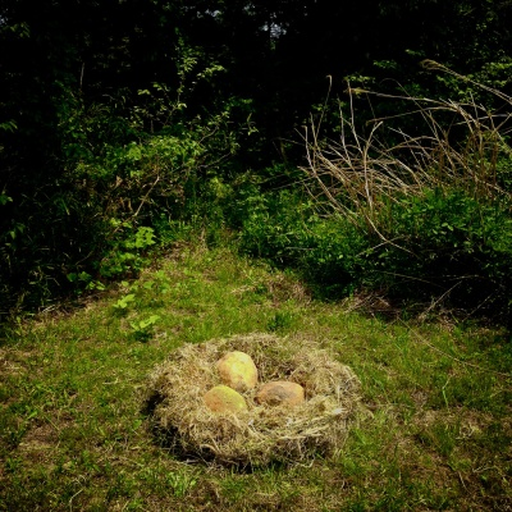}
    \end{minipage}
     \begin{minipage}{0.21\linewidth}
     \centering
        \includegraphics[width=0.99\linewidth]{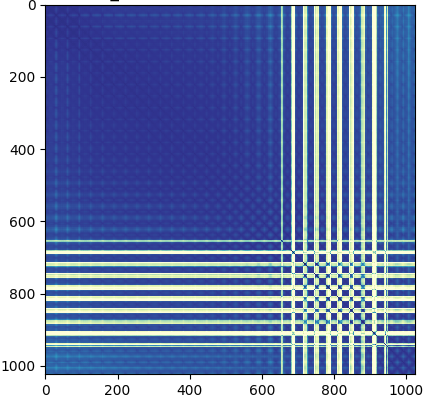}
    \end{minipage}
     \begin{minipage}{0.21\linewidth}
     \centering
        \includegraphics[width=0.99\linewidth]{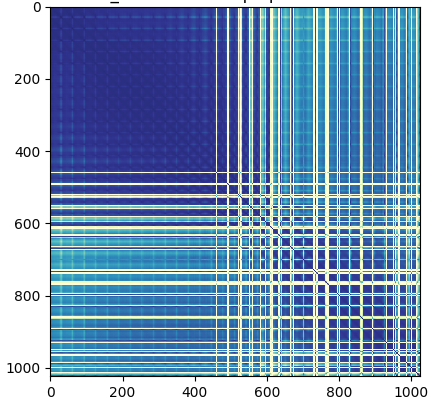}
    \end{minipage}
    \begin{minipage}{0.21\linewidth}
     \centering
        \includegraphics[width=0.99\linewidth]{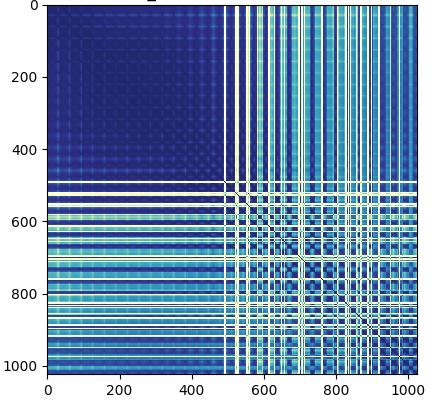}
    \end{minipage} 
    \begin{minipage}{0.04\linewidth}
     \centering          
    \includegraphics[width=0.99\linewidth]{images/ablation_study/scale_ablationstudy.png}
    \end{minipage}
    \begin{minipage}{0.21\linewidth}
     \centering
       ~
    \end{minipage}
    \begin{minipage}{0.21\linewidth}
     \centering
       ~
    \end{minipage}
    \begin{minipage}{0.21\linewidth}
     \centering
     \vspace{-0.15cm}
       (435.90, 0.72, 0.24)
    \end{minipage}
    \begin{minipage}{0.21\linewidth}
     \centering
     \vspace{-0.15cm}
       (420.68, 0.91, 0.58)
    \end{minipage}
     \begin{minipage}{0.21\linewidth}
     \centering
        \includegraphics[width=0.99\linewidth]{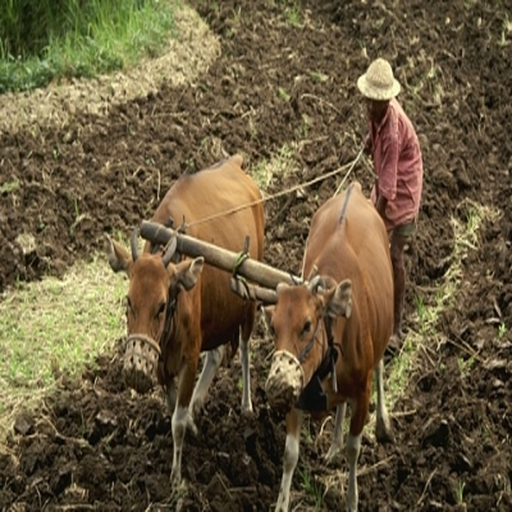}
    \end{minipage}
     \begin{minipage}{0.21\linewidth}
     \centering
        \includegraphics[width=0.99\linewidth]{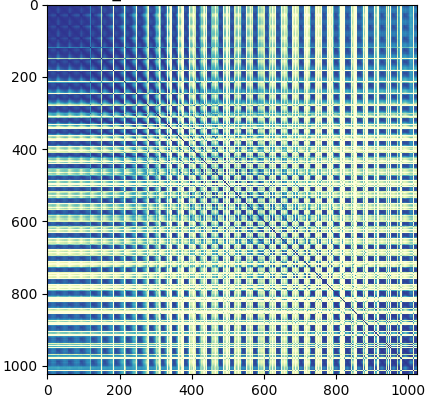}
    \end{minipage}
     \begin{minipage}{0.21\linewidth}
     \centering
        \includegraphics[width=0.99\linewidth]{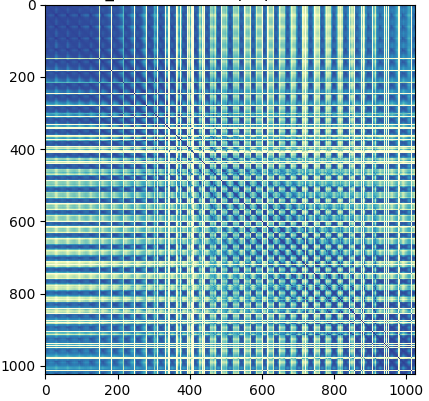}
    \end{minipage}
    \begin{minipage}{0.21\linewidth}
     \centering
        \includegraphics[width=0.99\linewidth]{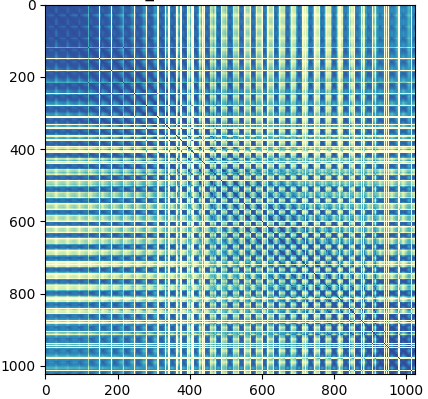}
    \end{minipage}
    \begin{minipage}{0.04\linewidth}
     \centering          
    \includegraphics[width=0.99\linewidth]{images/ablation_study/scale_ablationstudy.png}
    \end{minipage}
    \begin{minipage}{0.21\linewidth}
     \centering
       ~
    \end{minipage}
    \begin{minipage}{0.21\linewidth}
     \centering
       ~
    \end{minipage}
    \begin{minipage}{0.21\linewidth}
     \centering
    \vspace{-0.15cm}
       (207.59, 0.95, 0.81)
    \end{minipage}
    \begin{minipage}{0.21\linewidth}
     \centering
     \vspace{-0.15cm}
       (187.68, 0.96, 0.84)
    \end{minipage}
     \begin{minipage}{0.21\linewidth}
     \centering
        \includegraphics[width=0.99\linewidth]{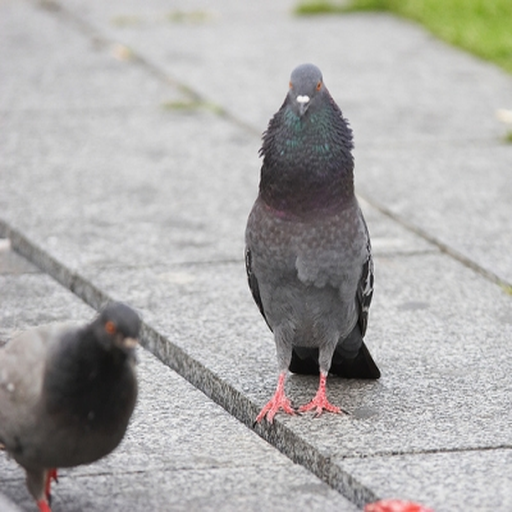}
    \end{minipage}
     \begin{minipage}{0.21\linewidth}
     \centering
        \includegraphics[width=0.99\linewidth]{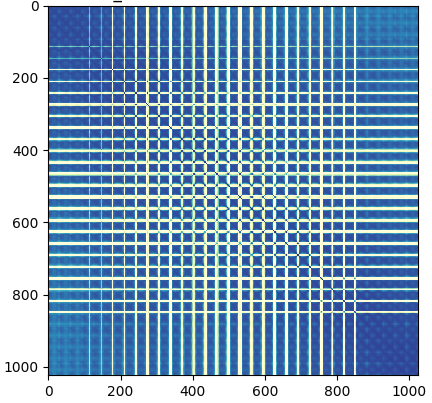}
    \end{minipage}
     \begin{minipage}{0.21\linewidth}
     \centering
        \includegraphics[width=0.99\linewidth]{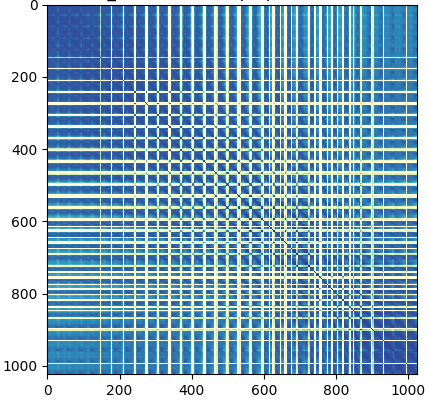}
    \end{minipage}
    \begin{minipage}{0.21\linewidth}
     \centering
        \includegraphics[width=0.99\linewidth]{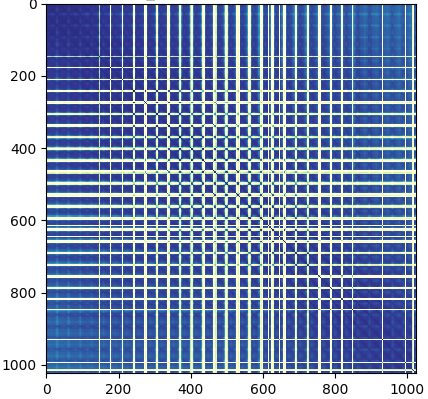}
    \end{minipage}
    \begin{minipage}{0.04\linewidth}
     \centering          
    \includegraphics[width=0.99\linewidth]{images/ablation_study/scale_ablationstudy.png}
    \end{minipage}
    \begin{minipage}{0.21\linewidth}
     \centering
       ~
    \end{minipage}
    \begin{minipage}{0.21\linewidth}
     \centering
       ~
    \end{minipage}
    \begin{minipage}{0.21\linewidth}
     \centering
     \vspace{-0.15cm}
       (205.52, 0.96, 0.75)
    \end{minipage}
    \begin{minipage}{0.21\linewidth}
     \centering
     \vspace{-0.15cm}
       (149.45, 0.99, 0.90)
    \end{minipage}
    \vspace{-0.3cm}
\caption{Values reported at bottom of DSM \cite{melas2022deep} and SimSAM (ours) method is Frobenius Norm, Accuracy and mIoU scores of randomly sampled image from ECSSD dataset. \textbf{Ablation Study-(I)} of main manuscript.}
\label{fig:fig21}
    % \begin{minipage}{0.95\linewidth}
    %  \centering
    %  \vspace{0.3cm}
    %  Values reported at bottom of each baseline method is Frobenius Norm, Accuracy and mIoU scores of randomly sampled image from ECSSD dataset. Ablation Study I of Paper. 
    \end{figure*}

\noindent \textbf{Image Segmentation.} We presented mIOU scores for evaluating the quality of the segmentation mask, as mentioned in Section~4 of the main paper. Our method outperformed the DSM method on ECSSD, DUTS and OMRON datasets. We randomly sampled ten images from ECSSD data and computed their Frobenius Norm, Accuracy and mIoU scores for this. We found that our method performed better on those individual images. See the Fig.~\ref{fig:fig20} and \ref{fig:fig21}. 


% Semantic Segmentation Outputs
\clearpage

%  Frobenius Norm, Accuracy and mIoU scores
\clearpage
\clearpage
% {\small
% \bibliographystyle{IEEEbib}
% \bibliography{strings,refs}
% }

\end{document}